\documentclass{article}

\usepackage[final]{neurips_2025}

\usepackage[utf8]{inputenc} 
\usepackage[T1]{fontenc}    
\usepackage{hyperref}       
\usepackage{url}            
\usepackage{booktabs}       
\usepackage{amsfonts}       
\usepackage{nicefrac}       
\usepackage{microtype}      
\usepackage{xcolor}         
\usepackage{amsmath}
\usepackage{multirow,enumitem}
\usepackage{multicol}
\usepackage{subcaption}
\usepackage{graphicx}
\usepackage{authblk}

\DeclareMathOperator*{\argmax}{arg\,max}

\title{AI Should Sense Better, Not Just Scale Bigger: Adaptive Sensing as a Paradigm Shift}

\newcommand{\firstauth}{\textsuperscript{*}}
\newcommand{\corresauth}{\textsuperscript{$\dagger$}}

\author[1]{Eunsu Baek\firstauth}
\author[1]{Keondo Park}
\author[2]{JeongGil Ko}
\author[1]{Min-hwan Oh}
\author[3]{Taesik Gong}
\author[1]{Hyung-Sin Kim\corresauth}

\affil[1]{Graduate School of Data Science, Seoul National University, Seoul, South Korea}
\affil[2]{School of Integrated Technology, Yonsei University, Seoul, South Korea}
\affil[3]{Department of Computer Science and Engineering, UNIST, Ulsan, South Korea}

\affil[1]{\texttt{\{beshu9407, gundo0102, minoh, hyungkim\}@snu.ac.kr}}
\affil[2]{\texttt{jeonggil.ko@yonsei.ac.kr}}
\affil[3]{\texttt{taesik.gong@unist.ac.kr}}

\begin{document}

\maketitle


\begingroup
    \renewcommand\thefootnote{}
    \footnotetext{\firstauth \href{https://edw2n.github.io}{edw2n.github.io}. \corresauth Corresponding author}
\endgroup

\vspace{-4ex}
\begin{abstract}
\vspace{-1.5ex}
Current AI advances largely rely on scaling neural models and expanding training datasets to achieve generalization and robustness. Despite notable successes, this paradigm incurs significant environmental, economic, and ethical costs, limiting sustainability and equitable access. 
Inspired by biological sensory systems, where adaptation occurs dynamically at the input (e.g., adjusting pupil size, refocusing vision)---we advocate for \textit{adaptive sensing} as a necessary and foundational shift. Adaptive sensing proactively modulates sensor parameters (e.g., exposure, sensitivity, multimodal configurations) at the input level, significantly mitigating covariate shifts and improving efficiency. Empirical evidence from recent studies demonstrates that adaptive sensing enables small models (e.g., EfficientNet-B0) to surpass substantially larger models (e.g., OpenCLIP-H) trained with significantly more data and compute. 
We (i) outline a roadmap for broadly integrating adaptive sensing into real-world applications spanning humanoid, healthcare, autonomous systems, agriculture, and environmental monitoring, (ii) critically assess technical and ethical integration challenges, and (iii) propose targeted research directions, such as standardized benchmarks, real-time adaptive algorithms, multimodal integration, and privacy-preserving methods. 
Collectively, these efforts aim to transition the AI community toward sustainable, robust, and equitable artificial intelligence systems.
\vspace{-4ex}
\end{abstract}    
\vspace{-1ex}
\section{Introduction}
\vspace{-1ex}

Early neural-network research drew inspiration from cortical neurons, implicitly framing intelligence as a predominantly brain-centered phenomenon~\cite{mcculloch1943logical, rosenblatt1958perceptron}. Correspondingly, recent advancements in artificial intelligence (AI) have predominantly relied on scaling model size and expanding training datasets~\cite{team2023gemini,achiam2023gpt,touvron2023llama2,Qwen2.5-VL}. Although effective, this \textit{model-centric paradigm} introduces significant and unsustainable challenges, such as massive computational demands that exacerbate environmental degradation~\cite{brown2020language,patterson2021carbon}, socioeconomic inequities due to resource concentration in a few well-funded institutions~\cite{bender2021dangers,bommasani2021opportunities}, and persistent failures in generalizing under real-world domain shifts~\cite{hendrycks2021many,koh2021wilds}.

In contrast, biological cognition is fundamentally embodied: the brain operates within an integrated system comprising sensory organs (e.g., vision, hearing, and touch) and musculature, which forms critical perception-action loops essential for survival and evolutionary success. For robust perception, human sensory systems dynamically adapt at the sensor level both before and during cortical processing. 
For instance, humans correct optical blur by wearing corrective lenses rather than retraining their neural circuits with thousands of out-of-focus images. Similarly, when facing low contrast or glare, humans instinctively squint, rapidly narrowing the pupil to sharpen depth-of-field and suppress stray light. 
Thus, intelligence emerges from the coordinated evolution and dynamic interplay between sensory organs, neural pathways, and adaptive behaviors, not solely from neural complexity alone~\cite{kandel2000principles, squire2012fundamental}.
%
%
This biological principle highlights that effective perception and generalization in AI depend \textbf{not only on neural capacity but also on dynamic, real-time sensor adaptation} throughout the entire perception pipeline.

However, current AI methodologies still primarily address distribution shift by \textit{scaling} neural architectures and datasets~\cite{oquab2023dinov2,cherti2023openclip,achiam2023gpt,radford2023robust}, leaving the sensing interface unchanged. 
A few existing sensor-aware strategies, such as active perception, which repositions sensors or robots to optimize viewpoints~\cite{bajcsy1988active,shi2023real}, and sensor-fusion budgeting, which schedules when and what modalities to activate~\cite{tilmon2023energy}, operate at the system or motion-planning level.
Post-hoc physics-based~\cite{agarwal2021simulation, planche2021physics} and physics-informed (e.g., PINNs, DSE)~\cite{raissi2019pinns,maier2018dse} sensor simulations cannot recover the coupled environment–sensor dynamics once analog signals are digitized, as the measurements have already been shaped by sensor configurations, leading to irreversible information loss. They neglect how raw analog signals are digitized in the first place to best serve the downstream model.
Historically, adaptive optics (e.g., radar CFAR~\cite{finn1966radar,rohling2007radar}, astronomical seeing correction~\cite{babcock1953astronomical}) 
were designed to enhance human perception or measurement fidelity, but were grounded in physical or human-centric criteria rather than how sensors adapt to model’s representation space.

%
Only recently have benchmarks begun isolating the impact of intrinsic sensor settings, such as exposure and gain, on recognition accuracy~\cite{baek2024unexplored,baek2025adaptive}. Inspired by these insights, the concept of \textbf{test-time input adaptation} has emerged: dynamically optimizing sensor parameters frame-by-frame to deliver model-friendly data. 
A first prototype, Lens~\cite{baek2025adaptive}, significantly outperforms traditional (human-oriented) auto-exposure methods, maintaining high classification accuracy despite a 50$\times$ reduction in model size. Preliminary evidence suggests even greater potential: with ideal sensor adaptation, a lightweight 5M-parameter EfficientNet-B0~\cite{manion2024effect} can surpass the 632M-parameter OpenCLIP-H~\cite{openclip} trained on 160$\times$ more 
training data~\cite{baek2024unexplored}. These early results highlight both the promise of adaptive sensing and the urgency for systematic exploration.

While initial empirical demonstrations come from image-classification tasks, adaptive sensing is broadly applicable across diverse domains. By shaping incoming photons, acoustic waves, or tactile forces \textit{at the sensor}, it boosts efficiency and robustness in fields such as medical diagnostics~\cite{pooch2020can, albadawy2018deep}, autonomous driving~\cite{sakaridis2018semantic}, surveillance~\cite{wang2003adaptive}, and environmental monitoring~\cite{akiva2022self,Marsocci_2023_CVPR}.
This capability is becoming critical as  AI increasingly transitions from artificial, controlled simulations to  \textbf{physical, embodied deployments}~\cite{paolo24embodid}. In real-world robots and wearable devices, identical models may receive data from lenses, CCDs, MEMS microphones, or tactile skins, each differing across vendor, firmware, and production batch. These subtle differences introduce domain shifts, either explicit or latent, that are difficult to address through model advancement alone.

Recent embodied-AI benchmarks in locomotion~\cite{sferrazza2024humanoidbench, li2023behavior}, household manipulation~\cite{srivastava2022behavior}, and interactive task execution~\cite{yang2025embodiedbench} exemplify an urgent reality, where the future AI must sense, reason, and interact from and with the real world within strict real-time and on-device constraints. 
Physical AI platforms from micro-drones to neural prostheses face inherent limits in size, energy, and thermal dissipation, 
suggesting limits to the traditional ``bigger model, bigger dataset'' paradigm. 
Adaptive sensing provides a \textbf{complementary solution}: by reshaping inputs directly at the source\textit{ (i.e., at the hardware level)}, it delivers model-friendly signals that enable smaller networks to thrive. Just as the human visual system couples rapid eye movements to visual cognition, next-generation AI systems must integrate closed-loop, real-time sensor control and preprocessing 
into their inference pipeline.

\textbf{Our Position.} 
Inspired by 
the principle in biological sensory systems, 
%
we argue that \textbf{AI research must transition away from an exclusively model-centric paradigm toward prioritizing adaptive, input-level optimization as a first-class concern.} Adaptive sensing is more than an incremental advance. Rather, it represents a critical, paradigm-level shift toward sustainable, equitable, and robust AI.  
Sections~\ref{sec:limit}-\ref{sec:directions} articulate key research directions, evaluation frameworks, and interdisciplinary opportunities essential to establish adaptive sensing as a foundational pillar of future AI.
\vspace{-0.5ex}

\vspace{-0.5ex}
\section{Current State of AI: Limitations of the Model-Centric Paradigm}\label{sec:limit}
\vspace{-1.5ex}

The dominant approach in modern AI emphasizes scaling models and expanding datasets to improve robustness and generalization~\cite{brown2020language,bommasani2021opportunities}. Although this model-centric paradigm has yielded notable performance gains, it faces fundamental limitations that increasingly threaten its long-term viability:  
\vspace{-1ex}
\begin{itemize} [leftmargin=*]
    \item \textbf{Environmental and Computational Cost:} 
    Modern AI models, particularly large-scale models such as GPT-3/4~\cite{brown2020language,achiam2023gpt}, and advanced multimodal models~\cite{team2023gemini,touvron2023llama,touvron2023llama2}, require massive computational resources for training~\cite{cottier2024rising}. Training GPT-3 alone requires approximately 1.287 GWh of electricity, emitting roughly 552 tons of CO$_2$, comparable to driving a passenger vehicle for more than one million kilometers~\cite{patterson2021carbon, strubell2020energy}. The ongoing trend toward even larger models~\cite{kaplan2020scaling,hoffmann2022training} exponentially escalates these environmental costs, posing significant sustainability challenges.

    \item \textbf{Accessibility and Inequity:} 
    The heavy computational requirements of training and deploying state-of-the-art models (e.g., GPT variants, CLIP, and advanced multimodal transformers) restrict access primarily to organizations with substantial financial and computational resources~\cite{bommasani2021opportunities}. This centralization widens the global digital divide by disproportionately benefiting wealthy institutions and regions while restricting innovation and participation from smaller-scale researchers, startups, and economically disadvantaged groups. Consequently, the potential diversity of ideas and equitable distribution of AI benefits remain severely constrained.

    \item \textbf{Real-World Generalization Failures:} 
    AI models trained on massive yet static datasets frequently fail to generalize effectively when encountering novel or dynamically changing real-world conditions. Domain shifts--variations in environmental, sensor-specific, and task-specific contexts--are notoriously challenging for traditional data-centric approaches~\cite{datasetshift2009}. Existing robustness benchmarks inadequately capture the true complexity of real-world conditions~\cite{baek2024unexplored}. As a result, models trained under these benchmarks often exhibit significant performance drops when deployed in realistic settings, limiting their reliability in critical applications such as autonomous driving~\cite{sakaridis2018semantic}, medical diagnostics~\cite{pooch2020can, albadawy2018deep} and environmental monitoring~\cite{akiva2022self,Marsocci_2023_CVPR}.

    \item \textbf{Economic Sustainability and Scalability:} 
    The continuous increase in model complexity and training data volumes imposes a substantial financial burden~\cite{strubell2020energy, castro2024artificial}. The
    costs associated with advanced computational infrastructure, specialized hardware, and extensive data acquisition quickly become prohibitive. This economic barrier hampers the scalability and widespread adoption of advanced AI technologies, especially in resource-constrained settings or smaller-scale enterprises.

    \item \textbf{Ethical and Societal Concerns:} 
    Large-scale models inherently risk amplifying biases present in extensive, complex training datasets~\cite{hall2022systematic,kotek2023gender}.     
    As data volumes grow, auditing, identifying, and mitigating embedded biases becomes increasingly difficult~\cite{schwartz2022towards}. Without rigorous oversight, these biases can propagate societal harm, perpetuating stereotypes, inequities, and systemic prejudices~\cite{bender2021dangers}. Addressing these ethical implications through a model-centric lens alone remains insufficient, reinforcing the need for alternative paradigms.

\end{itemize}
\vspace{-1ex}
Collectively, these limitations emphasize the urgency of moving beyond a solely model-centric paradigm toward strategies that integrate smarter, adaptive sensing and context-aware optimization.

\vspace{-1ex}
\section{Adaptive Sensing as a Necessary Paradigm Shift}
\vspace{-1ex}

Overcoming the limitations of the model-centric paradigm requires treating sensor-level adaptability as a first-class design principle.
While biological sensors embed rapid, energy-proportional adaptation in hardware, modulating gain, bandwidth, and spatial resolution before neural inference begins, modern artificial sensors, such as cameras, microphones, and haptic arrays, \textbf{remain predominantly static} (Table~\ref{tab:comparison}). 
Bridging this gap through adaptive sensing at the input level would enable smaller, faster, and fairer AI models, providing a compelling path toward sustainable and equitable artificial intelligence.

\begin{table}[h]
    \centering    
    \vspace{-3ex}
    \caption{Comparison between adaptive human sensors and static artificial sensors.}
  \resizebox{\linewidth}{!}{
    \begin{tabular}{lll}
    \toprule
         \textbf{Modality} & \textbf{Human Sensor (Adaptive)} & \textbf{Artificial Sensor (Static)} \\
         \midrule
         \textbf{Vision} &  Pupil diameter 2–8 mm ($\sim$16$\times$ light gain) in < 200 ms. & Fixed or two-step aperture; ISO/shutter in coarse steps~\cite{gutschick2002should}. \\
         & Dark adaptation restores sensitivity; ciliary muscles refocus from 10 cm to infinity. & Autofocus 50–300 ms; quantum efficiency and CFA static. \\
         & Saccades (3–5 ms) redirect fovea before cortex~\cite{manion2024effect, webster2024designing}.    &  Saturation corrected only post-capture. \\
         \midrule
         \textbf{Hearing} & Active gain control (> 120 dB dynamic range) via outer-hair cells~\cite{lanting2013mechanisms}. &  60-90 dB dynamic range (MEMS mic) with fixed AGC~\cite{shah2019design}. \\
         & Real-time efferent feedback loop protects hearing from damage~\cite{moore2012introduction}. & Limited impulse protection.  \\
         & 3-D localization from binaural delay down to 10 µs. & 3D localization requires multi-element arrays and DSP. \\          
         \midrule
         \textbf{Touch} & Sensitivity of 0.3-500+ kPa via mosaic of mechanoreceptors. & Sensitivity of 1-100 kPa via capacitive or piezoresistive arrays. \\
         & Spatial acuity up to 0.5mm; temporal resolution $\sim$ 1kHz~\cite{van2016haptic, johansson2009coding}. & Taxel pitch ~1mm; sampling 100-500Hz~\cite{guo2024active}. \\
         & Soft, compliant skin spreads out pressure; nociceptors trigger pain reflexes. & Mechanically rigid surface with no pain feedback~\cite{jiang2025can}. \\
         \bottomrule
    \end{tabular}}
    \vspace{-2.5ex}
    \label{tab:comparison}
\end{table} 

\subsection{Early Evidence from Vision Tasks}
\vspace{-1ex}

\begin{figure}
    \centering
    \includegraphics[width=\linewidth, bb=0 0 1600 400]{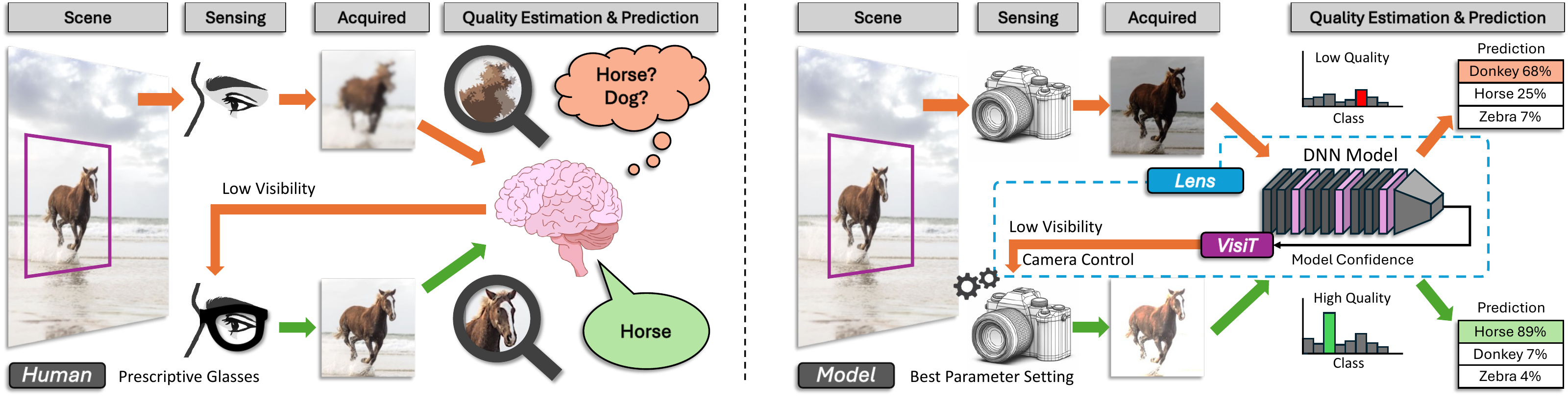}
    \vspace{-4ex}
    \caption{A representative adaptive sensing framework (Lens~\cite{baek2025adaptive}).}
    \label{fig:workflow-of-lens}
    \vspace{-3ex}
\end{figure}

Recent work on ImageNet-ES~\cite{baek2024unexplored} and ImageNet-ES-Diverse~\cite{baek2025adaptive} evaluates real-world covariate shifts induced by controlled sensor variation.  
Lens~\cite{baek2025adaptive} (Figure~\ref{fig:workflow-of-lens}) is the first model-friendly, test-time input adaptation framework to mitigate covariate shifts for image classification. It operates in a post-hoc, adaptive, and camera-agnostic sensor control manner, dynamically responding to scene characteristics based on VisiT scores to provide optimal image quality for neural networks. 
The key empirical findings are as follows, which underscore both the promise and the necessity of adaptive sensing methodologies: 
\vspace{-1ex}
\begin{itemize}[leftmargin=*]
\item \textbf{Accuracy gain:} Adaptive sensing significantly improves
accuracy up to 47.58\%p without model modification or maintain accuracy despite 50$\times$ model size difference. 

\item \textbf{Synergy:} Adaptive sensing synergistically integrates with model improvement techniques. 

\item \textbf{Specificity:} Adaptive sensing must be tailored in a model- and scene-specific manner.

\item \textbf{Human vs. Model optics:} High-quality images for model perception differ from those optimized for human perception.
\vspace{-1ex}
\end{itemize}

Beyond image classification, adaptive sensing has shown consistent benefits across \textit{diverse vision tasks}. In more complex settings such as object detection, segmentation, and remote photoplethysmography (rPPG), adaptive control of sensor parameters or viewpoints improves model robustness and accuracy under environmental shifts compared to non-adaptive configurations (e.g., fixed or auto-exposure, static viewpoints)~\cite{baek2025adaptive,madeye,odinaev2023rPPG}.
In 6D object-pose estimation—a multimodal task involving both depth and RGB channels—adaptive cross-modal coordination marks a new stage of adaptive sensing by jointly optimizing multiple sensing modalities~\cite{han2025senseshift6d}. It shows that the multi-modal control can achieve higher accuracy and stability than single-modal controls (e.g., RGB-only or depth-only) and surpasses factory defaults, while further enhancing robustness and data efficiency.
Together, these findings establish adaptive sensing as a scalable, complementary, and generalizable paradigm for robust perception across diverse sensing modalities and task conditions.

\subsection{Advantages for Real-World Agents and Deployment}
\vspace{-1ex}

\begin{table}[t]
\centering
\vspace{-0.5ex}
\caption{Broad real-world applicability of adaptive sensing.}
\fontsize{7.7}{9}\selectfont
\begin{tabular}{p{0.1\linewidth} p{0.84\linewidth}}
\toprule
\textbf{Domain} & \textbf{Role of Adaptive Sensing} \\
\midrule
\textbf{Humanoids} & 
Adaptive multimodal sensor adjustments—such as dynamically modulating visual, auditory, and tactile sensors—improve real-time balance, manipulation, and interaction quality in complex, unstructured environments. \\
\midrule
\textbf{Healthcare} & 
Dynamic adjustment of medical imaging parameters optimizes image quality based on patient-specific characteristics and contexts (e.g., dynamically adjusting MRI sequences to enhance diagnostic accuracy). \\
\midrule
\textbf{Self-driving} & 
Real-time optimization of camera and lidar parameters allows rapid adaptation to changing lighting and weather (e.g., fog, rain, or sudden brightness changes), enhancing perception robustness and road safety. \\
\midrule
\textbf{Agriculture} & 
Adaptive drone sensing systems dynamically adjust settings to capture high-quality data under varying conditions—such as crop type, growth stage, and environmental stress—for precise crop health monitoring. \\
\midrule
\textbf{Environment} & 
Adaptive sensing dynamically tunes sensor settings to capture accurate air and water quality data under diverse environmental conditions, improving monitoring accuracy and predictive modeling. \\
\bottomrule
\end{tabular}
\vspace{-3ex}
\normalsize
\label{tab:applicability}
\end{table}

Integrating adaptive sensing yields practical advantages in realistic operational settings:
\begin{itemize}[leftmargin=*]
    \vspace{-1ex}
    \item \textbf{Learning perspective:}
    Unlike simulation environments~\cite{mittal2023orbit,greff2022kubric,todorov2012mujoco}, real-world environments pose challenges due to sensor heterogeneity (cameras, microphones, haptic arrays) and unpredictable conditions (lighting, weather)~~\cite{li2023robustness,caesar2020nuscenes,jin2023development,humphries2020impact}. 
    Adaptive sensing enables agents to dynamically optimize sensor parameters (e.g., exposure, viewpoint), effectively reducing perceptual uncertainty. This targeted data acquisition approach enhances sample-efficient learning and robust generalization, especially under sparse reward conditions.

    \item \textbf{Engineering and Economic Viability:}
    Adaptive sensing reduces computational requirements compared to extensive model retraining, well-suited for resource-constrained and embedded systems. The improved data quality at the sensor level reduces infrastructure costs, enabling economically sustainable AI deployment at scale~\cite{van2021sustainable}.

    \item \textbf{Ethical and Societal Impact:} 
    By enabling targeted, context-aware data collection, adaptive sensing mitigates biases prevalent in large, static datasets~\cite{ruggeri2023multi, navigli2023biases, fraser2024examining}. This ensures fairer outcomes in sensitive applications, enhancing public trust and ethical integrity in AI systems.

    \item\textbf{Interdisciplinary Innovation.} 
    Adaptive sensing naturally encourages collaboration among computer scientists, engineers, ethicists, neuroscientists, and policymakers, fostering holistic solutions and comprehensive technological advancements.

    \item \textbf{Broad Domain Applicability:} 
    Adaptive sensing has the potential to practically impact diverse domains that utilize sensor data. Concrete scenarios are provided in Table~\ref{tab:applicability}.  
%
    



\end{itemize}
\vspace{-1ex}
In summary, adaptive sensing represents a necessary paradigm shift for smaller, greener, and fairer AI systems---turning ``sense better'' into ``learn less.''





\vspace{-1ex}
\section{Adaptive Sensing for Embodied AI: Obstacles and Outlook}
\vspace{-1ex}

\begin{figure}
    \captionsetup{labelformat=empty}
    \centering
    \begin{minipage}{0.51\textwidth}
        \begin{subfigure}[t]{1.0\textwidth}
            \centering
            \includegraphics[width=.85\linewidth, bb=0 0 611 351]{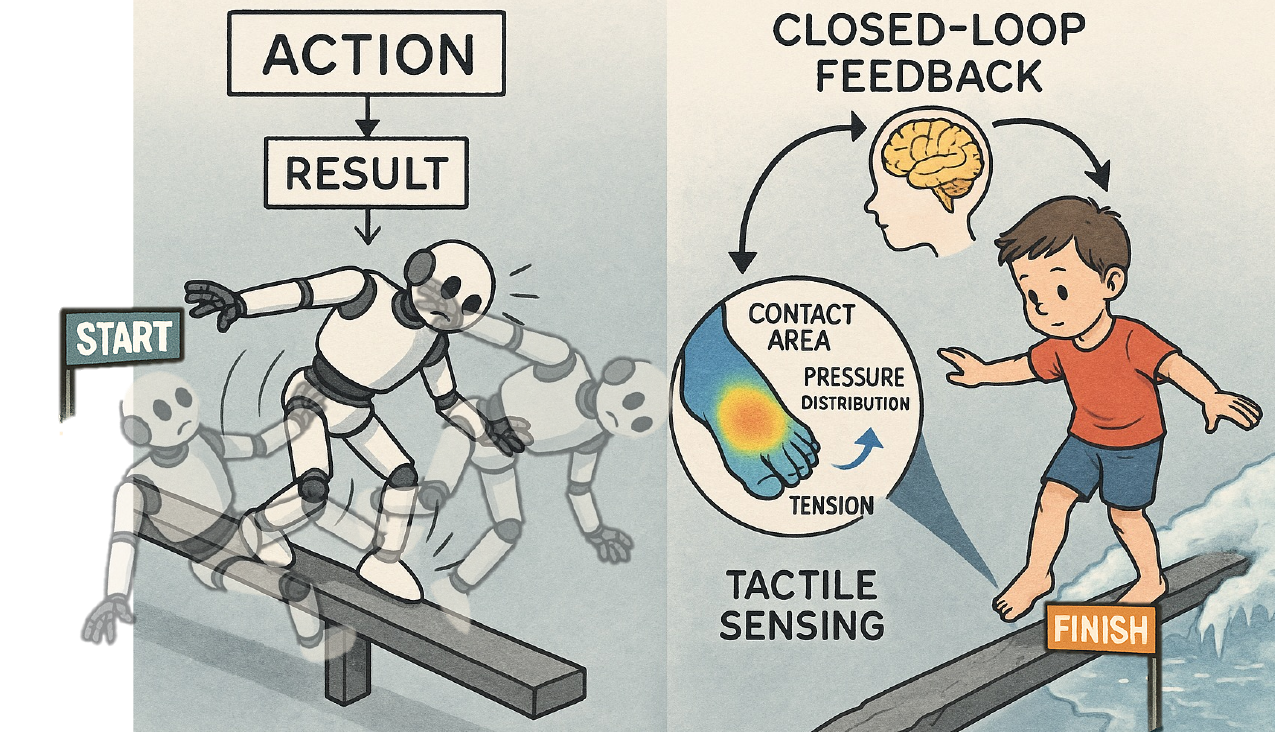}
            \vspace{-1ex}
            \caption{Physical AI (action-only) vs. child in a balancing task.}
            \label{fig:Vanila-vs-Closed-loop}
        \end{subfigure}        
        \begin{subfigure}[t]{1.0\textwidth}
            \centering
            \includegraphics[width=\linewidth, bb=0 0 532 464]{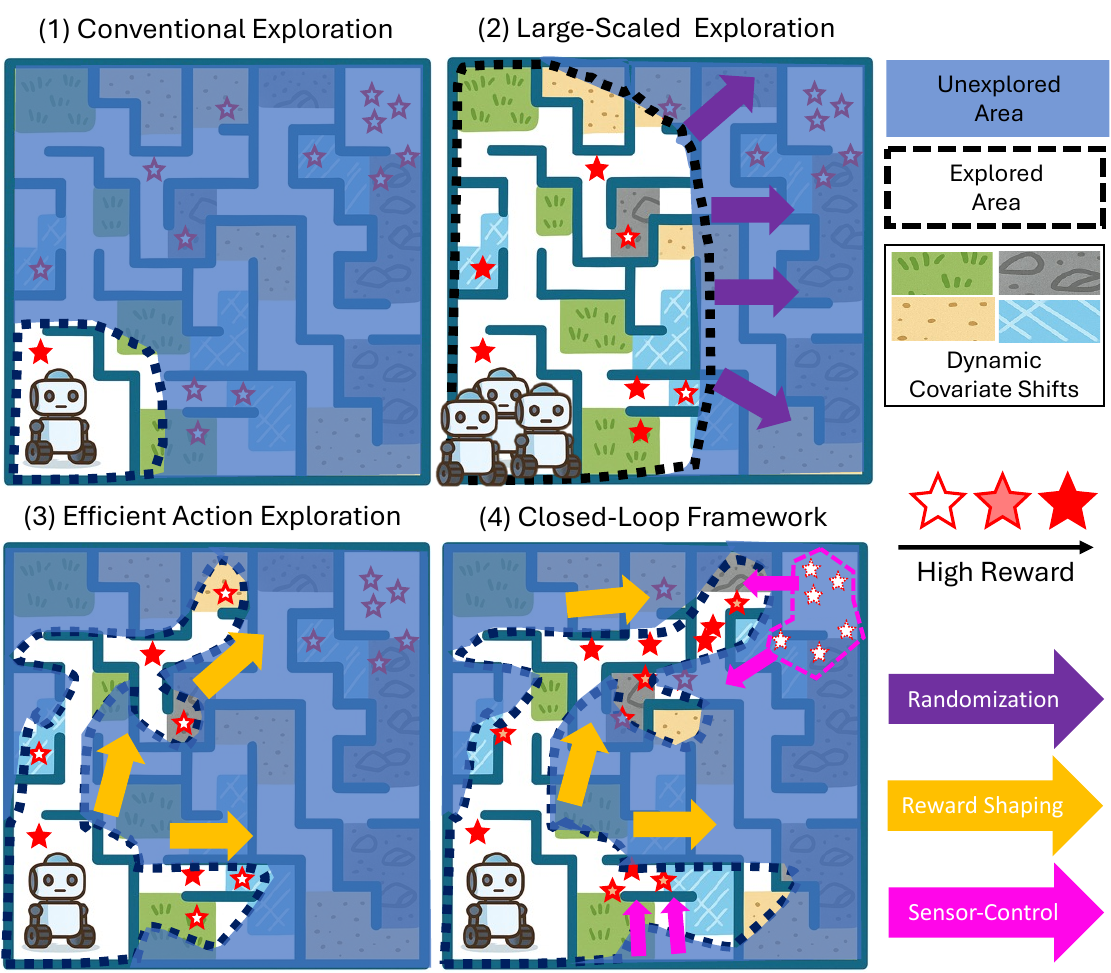}
            \vspace{-3ex}
            \caption{Exploration in dynamic, sparse-reward settings. (1–3): Motor-only learning vs. (4): Perception-aware learning.}
            \label{fig:sparse-rewards}
        \end{subfigure}    
    \end{minipage}     
    \hfill
    \begin{minipage}{0.44\textwidth}
        \begin{subfigure}{\textwidth}
        \centering
        \includegraphics[width=.995\linewidth, bb=0 0 333 552]{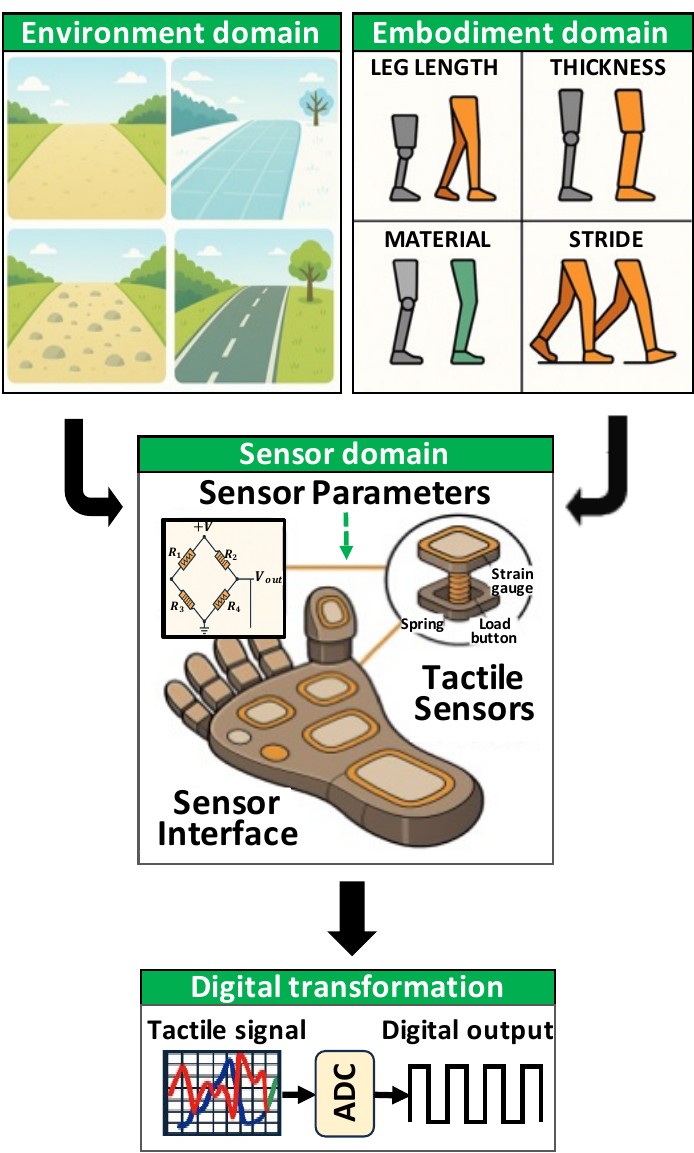}
        \caption{Illustration of covariate shift in embodied AI settings for a balancing task.}
        \label{fig:covaraiate-embodiai}
        \end{subfigure}        
    \end{minipage}
    \hfill  
    
    \vspace{-1ex}
    \captionsetup{labelformat=default}
    \caption{Why Closed-Loop Adaptive Sensing Framework is needed for Embodied AI Agents?}
    \label{fig:why-closed-loop}
    \vspace{-3ex}
\end{figure}

This section explores the deeper potential of adaptive sensing within embodied AI, expanding beyond the initial evaluation in static image classification tasks. 
Current embodied AI systems predominantly rely on \textit{action-centric training} without explicitly considering adaptive sensing, resulting in inefficient learning and limited adaptability~\cite{duan2022survey}. 
This limitation becomes particularly evident when comparing action-only embodied agents to humans, who seamlessly integrate sensory adaptation with motor actions in a closed-loop manner.
For instance, consider a balancing task (Fig.\,\ref{fig:Vanila-vs-Closed-loop}), which involves simultaneous covariate shifts~\cite{averly2023unified,yang2024generalized,baek2024unexplored} stemming from variations in environmental conditions, sensor interfaces and parameters, and agent morphologies (Fig.\,\ref{fig:covaraiate-embodiai}), and extremely sparse reward signals relative to the complexity of the multi-sensor, multi-modal exploration space (Fig.\,\ref{fig:sparse-rewards}).
Humans quickly achieve robust performance through efficient closed-loop sensory feedback, requiring only a few training trials and effortlessly adapting to new conditions. In contrast, embodied agents relying solely on motor actions significantly lag in both learning speed and real-world robustness.
%

\begin{figure}[t]
    \centering
    \begin{subfigure}{0.49\textwidth}
        \vspace{-1ex}
        \centering
        \includegraphics[width=\linewidth, bb=0 0 959 267]{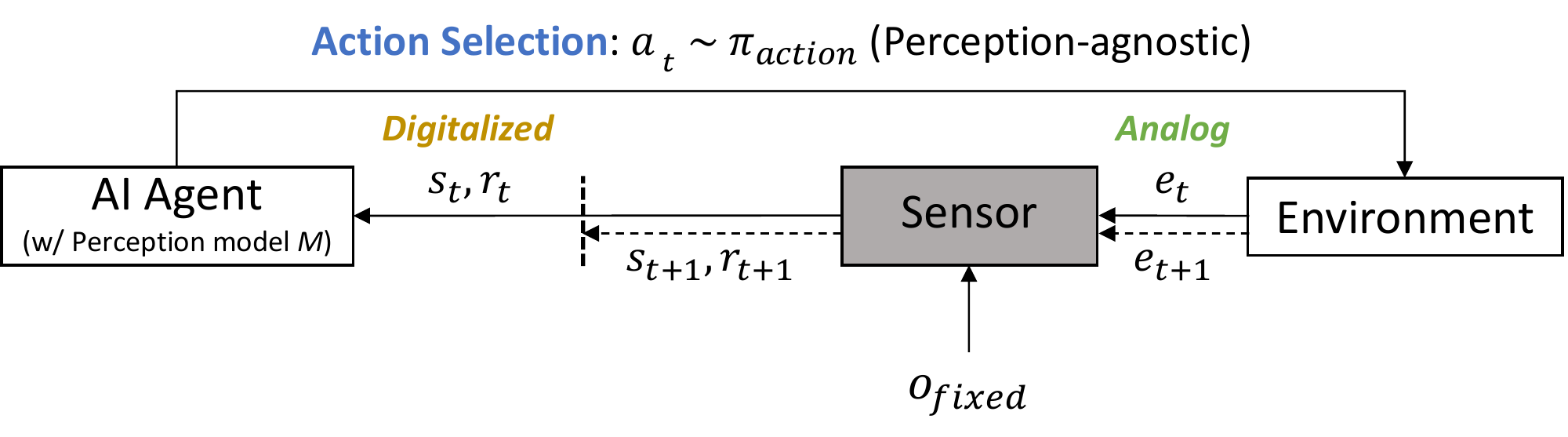}
        \vspace{-3ex}
        \caption{Conventional RL (MDP, No Adaptive Sensing).}
        \label{fig:framework-vanilla-rl}
    \end{subfigure}\hfill
    \begin{subfigure}{0.49\textwidth}
        \vspace{-1ex}
        \centering
        \includegraphics[width=\linewidth, bb=0 0 959 266]{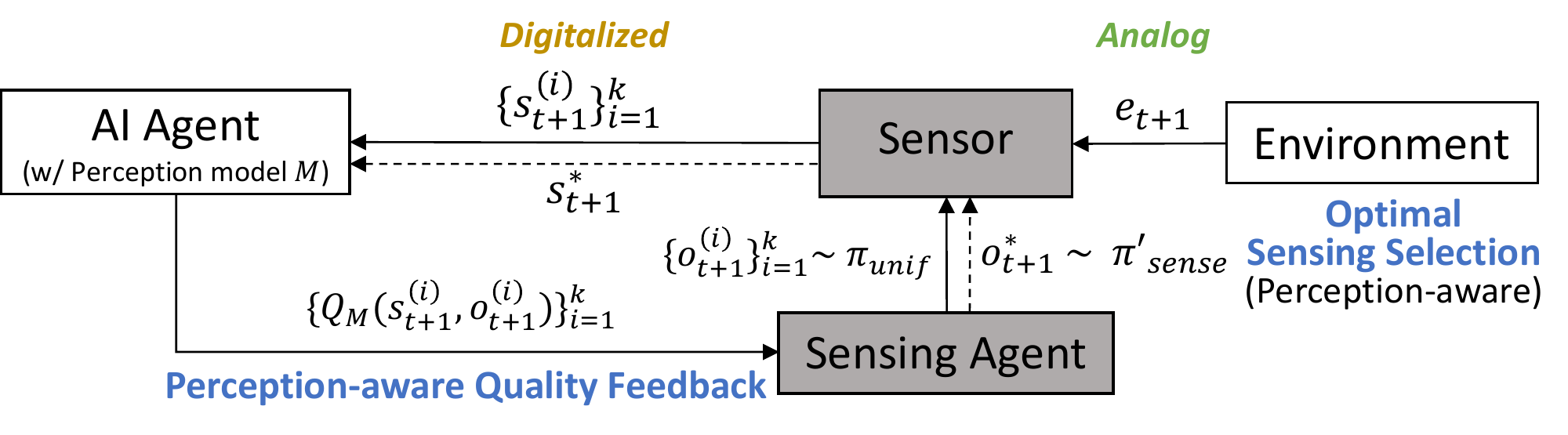}         
        \vspace{-3ex}
        \caption{Single-Shot-Perception (non-RL, Lens~\cite{baek2025adaptive}).}
        \label{fig:framework-single-shot}
    \end{subfigure}\hfill
    \begin{subfigure}{0.49\textwidth}        
        \centering
        \includegraphics[width=\linewidth, bb=0 0 959 290]{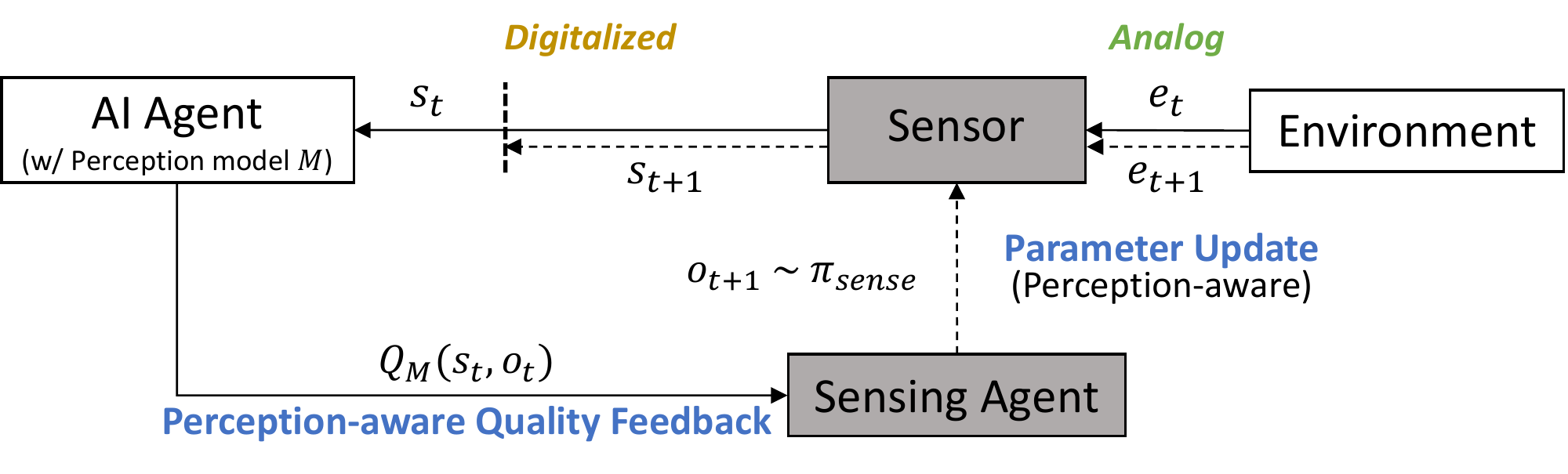}        
        \vspace{-3ex}
        \caption{Perception-Only.}
        \label{fig:framework-perception-only}
    \end{subfigure}\hfill
    \begin{subfigure}{0.49\textwidth}        
        \centering
        \includegraphics[width=\linewidth, bb=0 0 959 290]{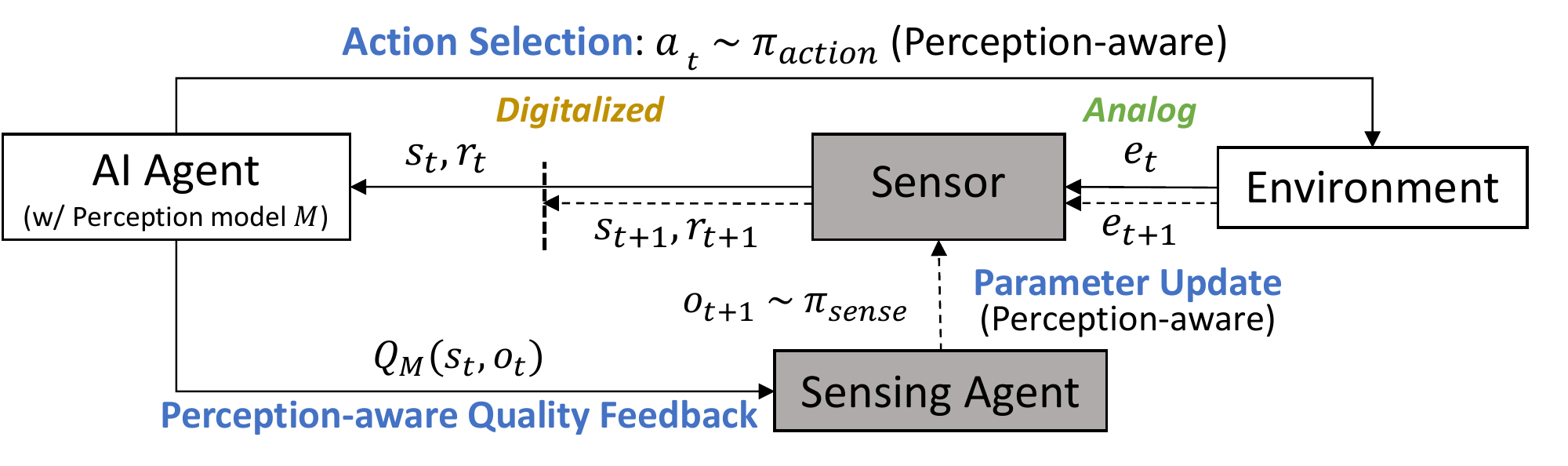}
        \vspace{-3ex}
        \caption{Sensorimotor.}
        \label{fig:framework-sensorimotor}
    \end{subfigure}
    \vspace{-1ex}
    \caption{Towards Closed-Loop Adaptive Sensing Framework for Embodied AI Agents.}
    \vspace{-3.5ex}
    \label{fig:Adaptive-Sensing-Framework}
\end{figure}

\textbf{Current Gaps in Adaptive Sensing Approaches:}   
Existing adaptive sensing methods, notably Lens~\cite{baek2025adaptive}, predominantly target single-shot perception tasks (Fig.~\ref{fig:framework-single-shot}), neglecting scenarios involving continuous interaction and sequential decision-making. These approaches fail to consider the crucial role of ongoing closed-loop feedback between sensing, perception, and action, which is essential for embodied agents operating within dynamic, real-world conditions. Moreover, current methods do not actively utilize model perception feedback to guide sensor parameter exploration. 
This limitation is especially critical in continuous, sparse-reward scenarios, where effective navigation of the sensor configuration space through closed-loop adaptive sensing is critical for robust and efficient learning. Addressing these gaps is imperative for successfully integrating adaptive sensing with continuous, action-driven learning paradigms in embodied AI.
\vspace{-1.5ex}
\section{Closed-Loop  Framework for Embodied AI Agents: Towards Humanoid}
\vspace{-1.5ex}

To fill the gaps and integrate adaptive sensing into embodied AI across both reinforcement learning (RL) and non-RL settings, we propose a principled \textbf{Adaptive Sensing Framework}. The framework is \textit{scalable} and covers \textit{essential settings} for embodied AI, starting from the most fundamental RL setting, a standard MDP formulation~\cite{puterman2014markov} as sequential decision-making. It also encompasses non-RL settings, drawing on single-shot control scenarios explored in early adaptive sensing studies, and progressively extends tocontinuous perception tasks, sensorimotor interactions, and ultimately multimodal embodied scenarios, all within a fully closed-loop adaptive structure. The ultimate goal is to establish a robust, adaptive AI pipeline capable of efficiently learning and reliably adapting to dynamic environments.

\vspace{-0.7ex}
\subsection{RL Agents with Sensing–Environment Interaction}
\vspace{-0.7ex}
To pave the way for closed-loop adaptive sensing in embodied AI, we first formalize two baseline setups—each reflecting that the agent’s state is observed by sensor measurements (via measurement function~\(f\)) from the interaction between sensor configuration and the environment:
\subsubsection{Common Notations}
\begin{description}[leftmargin=0pt]
\item [Basic Environment Components.] Given state space $\mathcal{S}$, action space $\mathcal{A}$, transition probability $P$, and reward function $R$, the environment response distribution $P_E(e_{t+1}\mid s_t,a_t)$ defines the probability of the next environmental state $e_{t+1}$ conditioned on the current state-action pair $(s_t,a_t)$, encompassing all potential covariate shifts arising within the environment domain as a consequence of the agent's actions. Similarly, the reward is generally sampled as:
$r_{t+1} \sim R(s_{t}, \cdot)$
and optionally may incorporate additional contextual information such as sensing or action history.
\item[Sensor Parameter Options Space $\mathcal{O}$ ($\subseteq \mathbb{R}^p$).]The set of all possible sensing settings for $p$ parameters.
An option $o_t \in \mathcal{O}$ represents settings of sensing parameters at time $t$. The option $o_{\text{fixed}} \in \mathcal{O}$ denotes a constant parameter configuration, typically used in non-adaptive (fixed) sensing settings.
\item [Sensor Measurement $f(e_t, o_t)$.] 
A sensor operation converting analog signals from the environment $e_t$ into a digital observation $s_t$ under sensor configuration $o_t$ at timestep $t$. This process inherently captures various covariate shifts resulting from sensor-environment interactions.

\item [Perception-Aware Quality Estimation $Q_M(s_t, \cdot)$.] 
A perception-based metric evaluating how easily an observation $s_t$ can be interpreted by a given perception model (or module) $M$. For example, if $M$ is an image classification model, the quality metric can be defined as the maximum confidence score (Lens~\cite{baek2025adaptive}): $Q_{M}(s_t) := \max(\text{softmax}(M(s_t))).$
Optionally, additional contextual information, such as sensing or action history, can be incorporated into this metric.
\item [Action Policy $\pi_{\text{action}}(a_t \mid s_t, \cdot)$.] 
General policy selecting the agent's action based on relevant states and contextual information such as action history (e.g., $\pi_{\text{action}}(a_t \mid s_{t}, a_{t-1})$).
\item [Sensing Policy $\pi_{\text{sense}}(o_{t+1} \mid s_t, \cdot)$.] 
General policy selecting the next sensing configuration conditioned on prior observations and contextual information such as sensing history, and perception-aware quality metrics 
(e.g., $\pi_{\text{sense}}(o_{t+1} \mid s_{t}, o_{t}, Q_{M})$).
\end{description}

\vspace{-0.5ex}
\textbf{Conventional RL (MDP) without Adaptive Sensing. (Fig.\,\ref{fig:framework-vanilla-rl})}
We represent the environment as a standard Markov Decision Process (MDP)~\cite{puterman2014markov}, \(\mathcal{M}=(\mathcal{S},\mathcal{A},P_{E},R)\)
assuming non-adaptive (i.e., fixed) sensor configurations. 
At each timestep \(t\), the agent executes:
\vspace{-1ex}
\begin{enumerate}[label=\textbf{\arabic*.}, leftmargin=*]
  \item \textbf{State Observation:} \quad $s_t \in \mathcal{S}.$
  \item \textbf{Action Selection:} \quad $a_t \sim \pi_{action}(a_t \mid s_t).$
  \item \textbf{Environment Response and Sensor Measurement (No Sensing Agent):}
\vspace{-0.5ex}
\begin{equation}\label{eq:data_acq_nonadaptive}
    e_{t+1} \sim P_E(e_{t+1} \mid s_t, a_t), \quad s_{t+1} \sim f(e_{t+1},o_{fixed})
\end{equation}
  \item \textbf{Reward Collection:} \quad $r_{t+1} \sim R(s_{t}, a_t)$
\end{enumerate}
\vspace{-1ex}
\textbf{Single‐Shot Adaptive Sensing (Perception-Only, No Actions, (Fig.\,\ref{fig:framework-single-shot})).} 
%
We define a newly augmented stochastic process $\mathcal{P}:=(\mathcal{S},\mathcal{O},P'_E,Q_M)$
defined by
$P'_E$ which is the environment transition probability similar to $P_E$ defined in Eq.\eqref{eq:data_acq_nonadaptive}, but agnostic to actions.
This new stochastic process $\mathcal{P}$
captures ways in which sensing configurations are adaptively selected per-timestep for single-shot perception tasks (e.g., image classification tasks). 
Concretely, at timestep $t$, the agent performs:
\vspace{-1ex}

\begin{enumerate}[label=\textbf{\arabic*.}, leftmargin=*]
  
  \item \textbf{State Candidate Observation:}
     For a given candidate sensor configuration $o_{t+1}$ at timestep $t+1$ and the current state $s_t$, a candidate sensor measurement is sampled via the following approach:
    \begin{equation}\label{eq:data_acq_adaptive}
        e_{t+1} \sim P'_E(e_{t+1}\mid s_{t}), \quad s_{t+1} \sim f(e_{t+1}, o_{t+1}).
    \end{equation}
  
  \item \textbf{Perception-Aware Quality Estimation:} Evaluate the quality of the candidate measurement using the perception-aware metric, $Q_M(s_{t+1}, o_{t+1})$.
  
  \item \textbf{Candidate Sampling (Perception-Agnostic Exploration):} To choose a well-performing sensor configuration $o_{t+1}^{*}$ for observing a promising state $s_{t+1}^{*}$, we first sample $k$ candidate configurations, where $k$ is a hyperparameter controlling the overall processing time. These configurations form a set: $O_{t+1} := \{o_{t+1}^{(1)}, o_{t+1}^{(2)}, \dots, o_{t+1}^{(k)}\},$ sampled uniformly without replacement from a state-agnostic policy $\pi_{\text{unif}}$ over the sensor configuration space $\mathcal{O}$. For each sampled candidate $o_{t+1}^{(i)}$ ($i = 1, \dots, k$), we obtain the corresponding candidate state $s_{t+1}^{(i)}$ according to Eq.~\eqref{eq:data_acq_adaptive}. The set of candidate states is then defined as: $S_{t+1} := \{s_{t+1}^{(1)}, s_{t+1}^{(2)}, \dots, s_{t+1}^{(k)}\}.$
  
  \item \textbf{Optimal Sensing Selection (Perception-Aware Sensing Policy) and State Observation:} Finally, the perception-aware sensing policy $\pi'_\text{sense}$ selects the optimal sensing configuration $o_{t+1}^{*}$ among candidates based on the perception-aware quality metric $Q_M$:
    \(
  o_{t+1}^{*} 
= \argmax_{i \in \{1,\dots,k\}} Q_M(s_{t+1}^{(i)}, o_{t+1}^{(i)}) =: \pi'_\text{sense}(O_{t+1} \mid S_{t+1}, Q_M).
    \)

The agent then observes the perceptually optimal state using the chosen sensing configuration:
\(
s_{t+1}^{*} \sim f(e_{t+1}, o_{t+1}^{*}).
\)
\end{enumerate}


\vspace{-2ex}

\subsection{Single-Modal Continuous Perception Tasks}
\vspace{-0.5ex}
As shown in Fig.~\ref{fig:framework-perception-only}, we generalize the single-shot adaptive sensing scenario (Lens; Fig.~\ref{fig:framework-single-shot}) to a sequential, continuous reinforcement learning (RL) setting without explicit physical actions. We formalize this scenario as a Markov Decision Process (MDP), defined by \(\mathcal{M}=(\mathcal{S},\mathcal{O},P'_E,Q_M)\), where the agent's decisions exclusively involve adaptively selecting sensing configurations \(o_t \in \mathcal{O}\). At each timestep \(t\), the agent performs the following:

\vspace{-1ex}
\begin{enumerate}[label=\textbf{\arabic*.}, leftmargin=*]
  \item \textbf{State Observation:} \quad Observe current state \(s_t \in \mathcal{S}\).
  \item \textbf{Perception-Based Sensing Selection:} \quad Evaluate perception-aware quality metric \(Q_{M}(s_t, o_t)\) and select the next sensor configuration via policy \(\pi_{\text{sense}}\):
  \(
    o_{t+1} \sim \pi_{\text{sense}}(o_{t+1} \mid s_t, o_t, Q_M).
  \)
  \item \textbf{Environment Response and Sensor Measurement:}
  \[
    e_{t+1} \sim P'_E(e_{t+1} \mid s_t), \quad s_{t+1} \sim f(e_{t+1}, o_{t+1}).
  \]
\end{enumerate}



\vspace{-1ex}
In contrast to Lens, where sensing parameters were selected randomly or via simple heuristics, the sensing policy here explicitly maximizes the model-centric quality metric \(Q\). Drawing inspiration from how infants progressively refine gaze control based on continuous perceptual feedback ~\cite{yu2024active, miyazaki2014image}, our closed-loop adaptive sensing strategy promotes efficient, coherent sensor exploration, resulting in robust and improved performance in single-modal continuous perception tasks.

\vspace{-0.5ex}
\subsection{Continuous Sensorimotor Tasks}
\vspace{-1ex}
\textbf{Key Intuition: Closed-loop Feedback Learning in Humans.} 
In real-world sensorimotor tasks, agents must dynamically co-adapt sensing parameters and physical actions to maintain balance and locomotion across varied terrains. Humans instinctively—or using tools—modulate sensor gains (e.g., foot-pressure thresholds) and adjust actions (e.g., stride lengths) according to environmental conditions such as icy, rocky, or uphill surfaces. Through repeated sensorimotor experiences within this closed-loop feedback framework, perception continuously informs actions, and actions shape subsequent sensory inputs, enabling robust, adaptive behavior.

We formalize this scenario as a Markov Decision Process (MDP) defined by \(\mathcal{M}=(\mathcal{S}, \mathcal{A}, \mathcal{O}, P_E, Q_M)\), where the agent jointly selects sensing configurations \(o_t \in \mathcal{O}\) and physical actions \(a_t \in \mathcal{A}\). At each timestep \(t\), the agent executes:

\vspace{-1ex}
\begin{enumerate}[label=\textbf{\arabic*.}, leftmargin=*]
  \item \textbf{State Observation:} \quad Observe current state \(s_t \in \mathcal{S}\).  
  \item \textbf{Perception-Aware Action Selection:} \quad Select the next action via policy \(\pi_{\text{action}}\), conditioned on perception-aware quality feedback:
  \(
    a_t \sim \pi_{\text{action}}(a_t \mid s_t, o_t, a_{t-1}, Q_M).
  \)
  \item \textbf{Perception-Aware Sensing Selection:} \quad Select the next sensor configuration via policy \(\pi_{\text{sense}}\):  
  \[
    o_{t+1} \sim \pi_{\text{sense}}(o_{t+1} \mid s_t, o_t, Q_M).
  \]
  \item \textbf{Environment Response and Sensor Measurement:}  
  \[
    e_{t+1} \sim P_E(e_{t+1} \mid s_t, a_t), \quad s_{t+1} \sim f(e_{t+1}, o_{t+1}).
  \]
  \item \textbf{Reward Collection:} \quad Given a hyperparameter \(\lambda\), the agent balances task-oriented performance with sensing-quality feedback, enabling efficient joint learning of sensing and action policies:
  \(
    r_{t+1} \sim R(s_t, a_t, o_t) = R_{\text{task}}(s_t, a_t) + \lambda\, Q_M(s_t, o_t).
  \)
  where $R_{task}$is a task-specific reward function that quantifies the effectiveness or success of the agent's actions in achieving the intended physical objective (e.g., duration of maintained balance).
\end{enumerate}

\vspace{-0.5ex}
\textbf{Adaptation to Multi-modality (Sensors) Settings.} 
Beyond adapting sensor parameters within a single modality, agents may also dynamically allocate attention across multiple sensory modalities depending on the task context. For example, when standing still, weight shifts toward the forefoot increase reliance on toe pressure sensors, whereas a lateral push engages ankle proprioceptors more heavily to restore balance. Formally, at each timestep \(t\), we introduce a modality-weight vector \(w_t \in \mathbb{R}^N\) across \(N\) available sensory modalities (e.g., pressure, torque, IMU), typically normalized so that \(\sum_{n=1}^{N} w_t[n] = 1\). Extending the single-modality sensing policy \(\pi_{\text{sense}}\) and measurement function \(f\), we define the multi-modality sensing policy \(\pi_{\text{multi-sense}}\) and measurement function \(f_{\text{multi-sense}}\) as:

\(
(o_{t+1}, w_{t+1}) = \pi_{\text{multi-sense}}\bigl(o_{t+1}, w_{t+1} \mid s_{t}, o_{t}, w_{t}, Q_M\bigr), \quad
s_{t+1} = f_{\text{multi-sense}}\bigl(s_{t}, o_t, w_t\bigr).
\)

By adapting \(w_{t+1}\) in response to environmental perturbations—such as increasing reliance on ankle sensors when experiencing lateral instability—the agent effectively focuses its sensing resources on the most informative modalities. This mechanism mirrors human sensorimotor reflexes, promoting rapid, robust adaptation and recovery.

\textbf{Humanoid-Scale Multimodal Tasks under Sparse Rewards.} 
For challenging humanoid tasks—such as opening a bottle cap—agents typically receive a binary, sparse reward \( R_{\text{sparse}} \in \{0,1\} \) only upon successful task completion. To facilitate efficient exploration and learning under these sparse reward conditions, we introduce intermediate perception-aware, continuous sensing-quality metrics \(Q_i\). These metrics quantify intermediate progress or cross-modal sensor feedback, thus providing informative and dense guidance.

For instance, if visual alignment is uncertain while closing a bottle cap, tactile feedback indicating increased grip tightness can enhance the visual sensing-quality metric. Conversely, uncertain tactile sensing can be clarified by visual information regarding precise object positioning. Formally, such cross-modal or intermediate-quality metrics are defined as follows:
\[
Q_{\text{grip}}(s_{t}, a_{t-1}, o_t^{\text{tact}}), \quad
Q_{\text{vis}}(s_{t}, a_{t-1}, o_t^{\text{cam}}, o_t^{\text{tact}}),
\]
where \( Q_{\text{grip}} \) measures grip stability from tactile sensors, and \( Q_{\text{vis}} \) evaluates visual alignment accuracy informed by both visual and tactile sensor configurations.

These quality metrics can be integrated into a composite reward function:
\[
R_t = R_{\text{sparse}} 
    + \lambda_{\text{tact}}\,Q_{\text{grip}}(s_{t}, a_{t-1}, o_t^{\text{tact}}) 
    + \lambda_{\text{vis}}\,Q_{\text{vis}}(s_{t}, a_{t-1}, o_t^{\text{cam}}, o_t^{\text{tact}}),
\]
where hyperparameters \( \lambda_{\text{tact}} \) and \( \lambda_{\text{vis}} \) control the relative contribution of each modality. Each metric \(Q_i\) explicitly captures task-relevant aspects of sensor data quality, such as grip stability, visual accuracy, uncertainty reduction, or information gain. By leveraging these intermediate, modality-specific metrics, the agent receives rich, continuous feedback tailored to downstream objectives, effectively mitigating the exploration challenges posed by sparse environmental rewards.

\vspace{-0.5ex}
\section{Challenges and Counterarguments}
\vspace{-1.5ex}


While adaptive sensing offers transformative potential to reshape AI design and operation, its widespread adoption faces several critical challenges and counterarguments.
\vspace{-1ex}
\subsection{Challenges}
\vspace{-1ex}
\begin{itemize}[leftmargin=*]
    \item \textbf{Lack of Benchmarks:} Existing benchmarks~\cite{deng2009imagenet, lin2014microsoft, panayotov2015librispeech, gemmeke2017audio} rarely capture sensor and environmental variations. Recent adaptive-sensing benchmarks~\cite{baek2024unexplored,baek2025adaptive} remain narrowly focused on image classification tasks, limiting generalizability and broader methodological insights.  

    \item \textbf{Unspecified Objectives:} Existing adaptive sensing strategies rely on metrics such as model confidence or out-of-distribution (OOD) scores~\cite{baek2025adaptive}, which inadequately capture data-quality under realistic covariate shifts. This hampers accurate assessments and practical applicability. 

    \item \textbf{Complexity of Multi-modal Sensor Spaces:} Multi-modal sensors enlarge the perceptual and solution spaces, increasing learning complexity---especially under sparse or delayed reward conditions. 
     
    \item \textbf{Performance Trade-offs:} Adaptive sensing strategies may at first underperform large-scale, resource-intensive models and may face real-time or deployment bottlenecks in practice. Stakeholders may prioritize immediate maximal accuracy over long-term sustainability and efficiency.

    \item \textbf{Integration Barriers with Existing Frameworks:} Implementing adaptive sensing may increase complexity and require significant modifications to existing AI pipelines, software stacks, and hardware platforms~\cite{tensorflow2015-whitepaper,paszke2019pytorch,wolf2019huggingface,van2024big}, posing logistical, economic, and organizational challenges. 

    
    \item \textbf{Ethical and Privacy Concerns:} Real-time, context-aware sensor optimization introduces potential ethical risks related to privacy, particularly when operating in sensitive environments. 
    \vspace{-1ex}
\end{itemize}

\vspace{-1ex}
\subsection{Counterarguments}
\vspace{-1ex}
\begin{itemize}[leftmargin=*]
    \item \textbf{Proprietary Sensors \& Transparency:} Although many sensors are proprietary, adaptive sensing does not require internal hardware access; existing APIs for exposure or gain control are sufficient. As demand grows, vendors are likely to expand such interfaces without compromising intellectual property. Overall, adaptive sensing is more likely to \textit{increase}, rather than reduce, transparency. 

    \item \textbf{Alternative methods:} 
    Approaches such as efficient architectures and federated learning (FL), often proposed within the current model-centric paradigm to address generalization and computational challenges, cannot recover information lost at capture time or resolve environment–sensor shifts. They also introduce trade-offs—efficiency may reduce OOD robustness and fairness, while FL suffers from non-IID bias and privacy risks. Adaptive sensing complements these methods by operating upstream, allowing compact models to match or even surpass large-scale ones.

    \item \textbf{Task-dependent or rapidly changing configurations:} In single-task deployments, Lens selects from a small runtime candidate pool, so task dependence is not limiting. For multi-tasking, condition the pool on the active task with a conditional/hierarchical policy. Under dynamics, avoid per-frame updates by triggering switches on drift/performance proxies (e.g., confidence, calibration, quality drops), stabilize with hysteresis and switch-cost/latency budgets, and schedule changes via history-/resource-aware bandits or reinforcement learning.
    \item \textbf{When and Why Adaptive Sensing Excels:} Adaptive sensing outperforms domain adaptation, domain generalization, and test-time adaptation under \textit{covariate shifts}, as hardware-level control directly reshapes raw measurements and can preserves information that post-hoc model adaptation cannot recover (§1).
    In contrast, under semantic or content shifts (e.g., unseen classes), model-level adaptation remains more effective. \textbf{The two paradigms are complementary}, addressing orthogonal failure modes and achieving the best performance when combined.
    In \textit{resource-constrained} settings, adaptive sensing remains attractive for its efficiency, enabling small models to rival much larger ones (§3).
    Determining when to apply each approach remains an open question (§7), as current OOD and uncertainty detectors primarily flag semantic shifts, leaving reliable differentiation between covariate and semantic (or mixed) shifts underexplored.

    \item \textbf{Why Closed-Loop? (vs. Separated):} A standalone controller (as in Lens~\cite{baek2025adaptive}) is feasible and effective in low-dynamic environments where no action policy is involved (Fig.~\ref{fig:framework-single-shot}). However, in embodied AI settings characterized by dynamic covariate shifts and sparse rewards, an optimal sensing strategy must be carefully designed to provide richer feedback and mitigate reward sparsity, as it depends on both the model’s perception characteristics and task objectives. Closed-loop co-optimization aligns sensor control with the agent’s policy learning, stabilizing observations and improving both sample efficiency and robustness (§4–5).

\end{itemize}

\vspace{-2ex}
\section{Open Research Directions}\label{sec:directions}
\vspace{-1.5ex}

Addressing the identified challenges requires concerted investigation across the following strategic research avenues: 
\vspace{-1.5ex}

\begin{itemize}[leftmargin=*]
    \item \textbf{Standardized Benchmarks:} Develop comprehensive, realistic benchmarks that evaluate dynamic sensor-level adaptation for various tasks, environments, and modalities, and develop synthetic simulations informed by the real-world data, enabling fair, robust, and comparable assessments. 

    \item \textbf{Data-quality Metrics:} 
    Develop test-time evaluation metrics reflecting the model's perception of sensor data quality under covariate shifts, leveraging both aleatoric and epistemic uncertainty. This will enable effective guidance and informed decisions in adaptive sensing.

    \item \textbf{Algorithmic Innovation in Real-time Adaptation:} Advance efficient algorithms using reinforcement learning~\cite{sutton1998reinforcement}, adaptive control~\cite{landau2011adaptive}, and online learning~\cite{hoi2021online} to optimize sensor parameters in dynamic environments. Developing frameworks to explore and exploit sensor configuration spaces will enhance responsiveness and efficiency. Future directions also include addressing practical deployment challenges such as multi-sensor synchronization, single-sensor multitasking, and deadline-constrained adaptation, which are key to mitigating real-time bottlenecks.

    \item \textbf{Co-Development of Models and Sensor Strategies:} Develop tightly integrated approaches that simultaneously optimize AI model architectures and adaptive sensor parameters, leveraging mutual feedback for improved generalization and robustness. 

    \item \textbf{Multimodal and Language-Driven Adaptive Sensing:} Integrate language-based and multimodal context into sensor adaptation strategies, facilitating intuitive human-AI interaction and broader applicability across domains such as robotics, healthcare, and interactive systems.

    \item \textbf{Privacy-Preserving Sensor Optimization:} Investigate secure, lightweight, and privacy-aware adaptive sensing methods suitable for on-device deployment. Interdisciplinary collaboration involving AI experts, ethicists, and policymakers will be critical for maintaining public trust. 
\end{itemize}

\vspace{-1.5ex}
\section{Conclusion}
\vspace{-1.5ex}

In this paper, we have advocated for adaptive sensing as a critical paradigm shift for overcoming fundamental limitations to the prevailing model-centric approach in AI. Drawing inspiration from robust biological sensory systems, adaptive sensing provides a practical pathway toward environmentally sustainable, computationally efficient, ethically responsible, and equitably distributed AI capabilities. By actively optimizing sensor parameters at the input stage, adaptive sensing significantly alleviates the computational and economic burdens currently in large-scale model training and deployment.

We have identified critical gaps in existing methodologies, their limited consideration of continuous, closed-loop interactions essential for real-world embodied agents. Addressing these gaps through targeted research---such as  standardized adaptive benchmarks, real-time adaptation algorithms, privacy and ethical standards, and multimodal sensor integration---will be vital. There are urgent, interdisciplinary opportunities that the broader NeurIPS community is uniquely positioned to tackle.

Given the escalating environmental, economic, and societal pressures resulting from conventional AI scaling, embracing adaptive sensing is not simply advantageous; it is imperative. We call upon the NeurIPS community to prioritize this research agenda. Now is the moment for decisive collective action---by integrating adaptive sensing, we can ensure AI's trajectory aligns fundamentally with ecological responsibility, ethical integrity, and global equity.

\newpage

\section*{Acknowledgments}
This work was supported in part by Institute of Information communications Technology Planning Evaluation
(IITP) grant funded by the Korea government (MSIT) (RS-2025-02263754), in part by the National Research Foundation (NRF) of Korea
grant funded by the Korea government (MSIT) (No. RS-2023-00222663).

\bibliographystyle{plain}
\bibliography{main}

@String(ICASSP=	{ICASSP})

@String(AAAI = {AAAI})

@inproceedings{
baek2025adaptive,
title={Adaptive Camera Sensor for Vision Models},
author={Eunsu Baek and Sung-hwan Han and Taesik Gong and Hyung-Sin Kim},
booktitle={The Thirteenth International Conference on Learning Representations},
year={2025},
url={https://openreview.net/forum?id=He2FGdmsas}
}

@inproceedings{baek2024unexplored,
  title={Unexplored faces of robustness and out-of-distribution: Covariate shifts in environment and sensor domains},
  author={Baek, Eunsu and Park, Keondo and Kim, Jiyoon and Kim, Hyung-Sin},
  booktitle={Proceedings of the IEEE/CVF Conference on Computer Vision and Pattern Recognition},
  pages={22294--22303},
  year={2024}
}

@article{oquab2023dinov2,
  title={Dinov2: Learning robust visual features without supervision},
  author={Oquab, Maxime and Darcet, Timoth{\'e}e and Moutakanni, Th{\'e}o and Vo, Huy and Szafraniec, Marc and Khalidov, Vasil and Fernandez, Pierre and Haziza, Daniel and Massa, Francisco and El-Nouby, Alaaeldin and others},
  journal={arXiv preprint arXiv:2304.07193},
  year={2023}
}

@inproceedings{cherti2023openclip,
  title={Reproducible scaling laws for contrastive language-image learning},
  author={Cherti, Mehdi and Beaumont, Romain and Wightman, Ross and Wortsman, Mitchell and Ilharco, Gabriel and Gordon, Cade and Schuhmann, Christoph and Schmidt, Ludwig and Jitsev, Jenia},
  booktitle={Proceedings of the IEEE/CVF Conference on Computer Vision and Pattern Recognition},
  pages={2818--2829},
  year={2023}
}

@inproceedings{deng2009imagenet,
  title={Imagenet: A large-scale hierarchical image database},
  author={Deng, Jia and Dong, Wei and Socher, Richard and Li, Li-Jia and Li, Kai and Fei-Fei, Li},
  booktitle={2009 IEEE conference on computer vision and pattern recognition},
  pages={248--255},
  year={2009},
  organization={Ieee}
}

@article{mcculloch1943logical,
  title={A logical calculus of the ideas immanent in nervous activity},
  author={McCulloch, Warren S and Pitts, Walter},
  journal={The bulletin of mathematical biophysics},
  volume={5},
  pages={115--133},
  year={1943},
  publisher={Springer}
}

@article{rosenblatt1958perceptron,
  title={The perceptron: a probabilistic model for information storage and organization in the brain.},
  author={Rosenblatt, Frank},
  journal={Psychological review},
  volume={65},
  number={6},
  pages={386},
  year={1958},
  publisher={American Psychological Association}
}

@book{kandel2000principles,
  title={Principles of neural science},
  author={Kandel, Eric R and Schwartz, James H and Jessell, Thomas M and Siegelbaum, Steven and Hudspeth, A James and Mack, Sarah and others},
  volume={4},
  year={2000},
  publisher={McGraw-hill New York}
}

@book{squire2012fundamental,
  title={Fundamental neuroscience},
  author={Squire, Larry and Berg, Darwin and Bloom, Floyd E and Du Lac, Sascha and Ghosh, Anirvan and Spitzer, Nicholas C},
  year={2012},
  publisher={Academic press}
}

@article{bajcsy1988active,
  title={Active perception},
  author={Bajcsy, Ruzena},
  journal={Proceedings of the IEEE},
  volume={76},
  number={8},
  pages={966--1005},
  year={1988},
  publisher={IEEE}
}

@article{shi2023real,
  title={Real-time multi-modal active vision for object detection on UAVs equipped with limited field of view LiDAR and camera},
  author={Shi, Chuanbeibei and Lai, Ganghua and Yu, Yushu and Bellone, Mauro and Lippiello, Vincezo},
  journal={IEEE Robotics and Automation Letters},
  volume={8},
  number={10},
  pages={6571--6578},
  year={2023},
  publisher={IEEE}
}

@inproceedings{tilmon2023energy,
  title={Energy-efficient adaptive 3D sensing},
  author={Tilmon, Brevin and Sun, Zhanghao and Koppal, Sanjeev J and Wu, Yicheng and Evangelidis, Georgios and Zahreddine, Ramzi and Krishnan, Gurunandan and Ma, Sizhuo and Wang, Jian},
  booktitle={Proceedings of the IEEE/CVF Conference on Computer Vision and Pattern Recognition},
  pages={5054--5063},
  year={2023}
}

@inproceedings{odinaev2023rPPG,
  title={Optimizing camera exposure control settings for remote vital sign measurements in low-light environments},
  author={Odinaev, Ismoil and Chin, Jing Wei and Luo, Kin Ho and Ke, Zhang and So, Richard HY and Wong, Kwan Long},
  booktitle={Proceedings of the IEEE/CVF Conference on Computer Vision and Pattern Recognition},
  pages={6086--6093},
  year={2023}
}

@article{han2025senseshift6d,
  title={SenseShift6D: Multimodal RGB-D Benchmarking for Robust 6D Pose Estimation across Environment and Sensor Variations},
  author={Han, Yegyu and Yoon, Taegyoon and Woo, Dayeon and Kim, Sojeong and Kim, Hyung-Sin},
  journal={arXiv preprint arXiv:2507.05751},
  year={2025}
}

@article{raissi2019pinns,
  title={Physics-informed neural networks: A deep learning framework for solving forward and inverse problems involving nonlinear partial differential equations},
  author={Raissi, Maziar and Perdikaris, Paris and Karniadakis, George E},
  journal={Journal of Computational physics},
  volume={378},
  pages={686--707},
  year={2019},
  publisher={Elsevier}
}

@inproceedings{agarwal2021simulation,
  title={Simulation of vision-based tactile sensors using physics based rendering},
  author={Agarwal, Arpit and Man, Timothy and Yuan, Wenzhen},
  booktitle={2021 IEEE International Conference on Robotics and Automation (ICRA)},
  pages={1--7},
  year={2021},
  organization={IEEE}
}

@inproceedings{planche2021physics,
  title={Physics-based differentiable depth sensor simulation},
  author={Planche, Benjamin and Singh, Rajat Vikram},
  booktitle={Proceedings of the IEEE/CVF International Conference on Computer Vision},
  pages={14387--14397},
  year={2021}
}

@inproceedings{maier2018dse,
  title={Deep scatter estimation (DSE): feasibility of using a deep convolutional neural network for real-time x-ray scatter prediction in cone-beam CT},
  author={Maier, Joscha and Berker, Yannick and Sawall, Stefan and Kachelrie{\ss}, Marc},
  booktitle={Medical imaging 2018: physics of medical imaging},
  volume={10573},
  pages={393--398},
  year={2018},
  organization={SPIE}
}

@inproceedings {madeye,
author = {Mike Wong and Murali Ramanujam and Guha Balakrishnan and Ravi Netravali},
title = {{MadEye}: Boosting Live Video Analytics Accuracy with Adaptive Camera Configurations},
booktitle = {21st USENIX Symposium on Networked Systems Design and Implementation (NSDI 24)},
year = {2024},
isbn = {978-1-939133-39-7},
address = {Santa Clara, CA},
pages = {549--568},
url = {https://www.usenix.org/conference/nsdi24/presentation/wong},
publisher = {USENIX Association},
month = apr
}

@article{babcock1953astronomical,
  title={The possibility of compensating astronomical seeing},
  author={Babcock, Horace W},
  journal={Publications of the Astronomical Society of the Pacific},
  volume={65},
  number={386},
  pages={229--236},
  year={1953},
  publisher={JSTOR}
}

@inproceedings{finn1966radar,
  title={Adaptive detection in clutter},
  author={Finn, Harold M},
  booktitle={Fifth Symposium on Adaptive Processes},
  pages={562--567},
  year={1966},
  organization={IEEE}
}

@article{rohling2007radar,
  title={Radar CFAR thresholding in clutter and multiple target situations},
  author={Rohling, Hermann},
  journal={IEEE transactions on aerospace and electronic systems},
  number={4},
  pages={608--621},
  year={2007},
  publisher={IEEE}
}

@inproceedings{strubell2020energy,
  title={Energy and policy considerations for modern deep learning research},
  author={Strubell, Emma and Ganesh, Ananya and McCallum, Andrew},
  booktitle={Proceedings of the AAAI conference on artificial intelligence},
  volume={34},
  number={09},
  pages={13693--13696},
  year={2020}
}

@article{patterson2021carbon,
  title={Carbon emissions and large neural network training},
  author={Patterson, David and Gonzalez, Joseph and Le, Quoc and Liang, Chen and Munguia, Lluis-Miquel and Rothchild, Daniel and So, David and Texier, Maud and Dean, Jeff},
  journal={arXiv preprint arXiv:2104.10350},
  year={2021}
}

@InProceedings{paolo24embodid,
  title = 	 {Position: A Call for Embodied {AI}},
  author =       {Paolo, Giuseppe and Gonzalez-Billandon, Jonas and K\'{e}gl, Bal\'{a}zs},
  booktitle = 	 {Proceedings of the 41st International Conference on Machine Learning},
  pages = 	 {39493--39508},
  year = 	 {2024},
  editor = 	 {Salakhutdinov, Ruslan and Kolter, Zico and Heller, Katherine and Weller, Adrian and Oliver, Nuria and Scarlett, Jonathan and Berkenkamp, Felix},
  volume = 	 {235},
  series = 	 {Proceedings of Machine Learning Research},
  month = 	 {21--27 Jul},
  publisher =    {PMLR},
  url = 	 {https://proceedings.mlr.press/v235/paolo24a.html}
}

@article{brown2020language, 
  title={Language models are few-shot learners},
  author={Brown, Tom and Mann, Benjamin and Ryder, Nick and Subbiah, Melanie and Kaplan, Jared D and Dhariwal, Prafulla and Neelakantan, Arvind and Shyam, Pranav and Sastry, Girish and Askell, Amanda and others},
  journal={Advances in neural information processing systems},
  volume={33},
  pages={1877--1901},
  year={2020}
}

@article{achiam2023gpt,
  title={Gpt-4 technical report},
  author={Achiam, Josh and Adler, Steven and Agarwal, Sandhini and Ahmad, Lama and Akkaya, Ilge and Aleman, Florencia Leoni and Almeida, Diogo and Altenschmidt, Janko and Altman, Sam and Anadkat, Shyamal and others},
  journal={arXiv preprint arXiv:2303.08774},
  year={2023}
}

@article{team2023gemini,
  title={Gemini: a family of highly capable multimodal models},
  author={Team, Gemini and Anil, Rohan and Borgeaud, Sebastian and Alayrac, Jean-Baptiste and Yu, Jiahui and Soricut, Radu and Schalkwyk, Johan and Dai, Andrew M and Hauth, Anja and Millican, Katie and others},
  journal={arXiv preprint arXiv:2312.11805},
  year={2023}
}

@article{touvron2023llama,
  title={Llama: Open and efficient foundation language models},
  author={Touvron, Hugo and Lavril, Thibaut and Izacard, Gautier and Martinet, Xavier and Lachaux, Marie-Anne and Lacroix, Timoth{\'e}e and Rozi{\`e}re, Baptiste and Goyal, Naman and Hambro, Eric and Azhar, Faisal and others},
  journal={arXiv preprint arXiv:2302.13971},
  year={2023}
}

@article{touvron2023llama2,
  title={Llama 2: Open foundation and fine-tuned chat models},
  author={Touvron, Hugo and Martin, Louis and Stone, Kevin and Albert, Peter and Almahairi, Amjad and Babaei, Yasmine and Bashlykov, Nikolay and Batra, Soumya and Bhargava, Prajjwal and Bhosale, Shruti and others},
  journal={arXiv preprint arXiv:2307.09288},
  year={2023}
}

@article{Qwen2.5-VL,
  title={Qwen2.5-VL Technical Report},
  author={Bai, Shuai and Chen, Keqin and Liu, Xuejing and Wang, Jialin and Ge, Wenbin and Song, Sibo and Dang, Kai and Wang, Peng and Wang, Shijie and Tang, Jun and Zhong, Humen and Zhu, Yuanzhi and Yang, Mingkun and Li, Zhaohai and Wan, Jianqiang and Wang, Pengfei and Ding, Wei and Fu, Zheren and Xu, Yiheng and Ye, Jiabo and Zhang, Xi and Xie, Tianbao and Cheng, Zesen and Zhang, Hang and Yang, Zhibo and Xu, Haiyang and Lin, Junyang},
  journal={arXiv preprint arXiv:2502.13923},
  year={2025}
}

@article{cottier2024rising,
  title={The rising costs of training frontier AI models},
  author={Cottier, Ben and Rahman, Robi and Fattorini, Loredana and Maslej, Nestor and Besiroglu, Tamay and Owen, David},
  journal={arXiv preprint arXiv:2405.21015},
  year={2024}
}

@inproceedings{radford2023robust,
  title={Robust speech recognition via large-scale weak supervision},
  author={Radford, Alec and Kim, Jong Wook and Xu, Tao and Brockman, Greg and McLeavey, Christine and Sutskever, Ilya},
  booktitle={International conference on machine learning},
  pages={28492--28518},
  year={2023},
  organization={PMLR}
}

@article{kaplan2020scaling,
  title={Scaling laws for neural language models},
  author={Kaplan, Jared and McCandlish, Sam and Henighan, Tom and Brown, Tom B and Chess, Benjamin and Child, Rewon and Gray, Scott and Radford, Alec and Wu, Jeffrey and Amodei, Dario},
  journal={arXiv preprint arXiv:2001.08361},
  year={2020}
}

@article{hoffmann2022training,
  title={Training compute-optimal large language models},
  author={Hoffmann, Jordan and Borgeaud, Sebastian and Mensch, Arthur and Buchatskaya, Elena and Cai, Trevor and Rutherford, Eliza and Casas, Diego de Las and Hendricks, Lisa Anne and Welbl, Johannes and Clark, Aidan and others},
  journal={arXiv preprint arXiv:2203.15556},
  year={2022}
}

@inproceedings{bender2021dangers,
  title={On the dangers of stochastic parrots: Can language models be too big?},
  author={Bender, Emily M and Gebru, Timnit and McMillan-Major, Angelina and Shmitchell, Shmargaret},
  booktitle={Proceedings of the 2021 ACM conference on fairness, accountability, and transparency},
  pages={610--623},
  year={2021}
}

@article{bommasani2021opportunities,
  title={On the opportunities and risks of foundation models},
  author={Bommasani, Rishi and Hudson, Drew A and Adeli, Ehsan and Altman, Russ and Arora, Simran and von Arx, Sydney and Bernstein, Michael S and Bohg, Jeannette and Bosselut, Antoine and Brunskill, Emma and others},
  journal={arXiv preprint arXiv:2108.07258},
  year={2021}
}

@inproceedings{hendrycks2021many,
  title={The many faces of robustness: A critical analysis of out-of-distribution generalization},
  author={Hendrycks, Dan and Basart, Steven and Mu, Norman and Kadavath, Saurav and Wang, Frank and Dorundo, Evan and Desai, Rahul and Zhu, Tyler and Parajuli, Samyak and Guo, Mike and others},
  booktitle={Proceedings of the IEEE/CVF international conference on computer vision},
  pages={8340--8349},
  year={2021}
}

@article{sferrazza2024humanoidbench,
  title={Humanoidbench: Simulated humanoid benchmark for whole-body locomotion and manipulation},
  author={Sferrazza, Carmelo and Huang, Dun-Ming and Lin, Xingyu and Lee, Youngwoon and Abbeel, Pieter},
  journal={arXiv preprint arXiv:2403.10506},
  year={2024}
}

@inproceedings{li2023behavior,
  title={Behavior-1k: A benchmark for embodied ai with 1,000 everyday activities and realistic simulation},
  author={Li, Chengshu and Zhang, Ruohan and Wong, Josiah and Gokmen, Cem and Srivastava, Sanjana and Mart{\'\i}n-Mart{\'\i}n, Roberto and Wang, Chen and Levine, Gabrael and Lingelbach, Michael and Sun, Jiankai and others},
  booktitle={Conference on Robot Learning},
  pages={80--93},
  year={2023},
  organization={PMLR}
}

@inproceedings{srivastava2022behavior,
  title={Behavior: Benchmark for everyday household activities in virtual, interactive, and ecological environments},
  author={Srivastava, Sanjana and Li, Chengshu and Lingelbach, Michael and Mart{\'\i}n-Mart{\'\i}n, Roberto and Xia, Fei and Vainio, Kent Elliott and Lian, Zheng and Gokmen, Cem and Buch, Shyamal and Liu, Karen and others},
  booktitle={Conference on robot learning},
  pages={477--490},
  year={2022},
  organization={PMLR}
}

@article{yang2025embodiedbench,
  title={EmbodiedBench: Comprehensive Benchmarking Multi-modal Large Language Models for Vision-Driven Embodied Agents},
  author={Yang, Rui and Chen, Hanyang and Zhang, Junyu and Zhao, Mark and Qian, Cheng and Wang, Kangrui and Wang, Qineng and Koripella, Teja Venkat and Movahedi, Marziyeh and Li, Manling and others},
  journal={arXiv preprint arXiv:2502.09560},
  year={2025}
}

@inproceedings{koh2021wilds,
  title={Wilds: A benchmark of in-the-wild distribution shifts},
  author={Koh, Pang Wei and Sagawa, Shiori and Marklund, Henrik and Xie, Sang Michael and Zhang, Marvin and Balsubramani, Akshay and Hu, Weihua and Yasunaga, Michihiro and Phillips, Richard Lanas and Gao, Irena and others},
  booktitle={International conference on machine learning},
  pages={5637--5664},
  year={2021},
  organization={PMLR}
}

@book{datasetshift2009,
author = {Quionero-Candela, Joaquin and Sugiyama, Masashi and Schwaighofer, Anton and Lawrence, Neil D.},
title = {Dataset Shift in Machine Learning},
year = {2009},
isbn = {0262170051},
publisher = {The MIT Press}
}

@article{sakaridis2018semantic,
  title={Semantic foggy scene understanding with synthetic data},
  author={Sakaridis, Christos and Dai, Dengxin and Van Gool, Luc},
  journal={International Journal of Computer Vision},
  volume={126},
  pages={973--992},
  year={2018},
  publisher={Springer}
}

@inproceedings{pooch2020can,
  title={Can we trust deep learning based diagnosis? the impact of domain shift in chest radiograph classification},
  author={Pooch, Eduardo HP and Ballester, Pedro and Barros, Rodrigo C},
  booktitle={Thoracic Image Analysis: Second International Workshop, TIA 2020, Held in Conjunction with MICCAI 2020, Lima, Peru, October 8, 2020, Proceedings 2},
  pages={74--83},
  year={2020},
  organization={Springer}
}

@article{albadawy2018deep,
  title={Deep learning for segmentation of brain tumors: Impact of cross-institutional training and testing},
  author={AlBadawy, Ehab A and Saha, Ashirbani and Mazurowski, Maciej A},
  journal={Medical physics},
  volume={45},
  number={3},
  pages={1150--1158},
  year={2018},
  publisher={Wiley Online Library}
}

@inproceedings{akiva2022self,
  title={Self-supervised material and texture representation learning for remote sensing tasks},
  author={Akiva, Peri and Purri, Matthew and Leotta, Matthew},
  booktitle={Proceedings of the IEEE/CVF Conference on Computer Vision and Pattern Recognition},
  pages={8203--8215},
  year={2022}
}

@InProceedings{Marsocci_2023_CVPR,
    author    = {Marsocci, Valerio and Gonthier, Nicolas and Garioud, Anatol and Scardapane, Simone and Mallet, Cl\'ement},
    title     = {GeoMultiTaskNet: Remote Sensing Unsupervised Domain Adaptation Using Geographical Coordinates},
    booktitle = {Proceedings of the IEEE/CVF Conference on Computer Vision and Pattern Recognition Workshops},
    month     = {June},
    year      = {2023},
    pages     = {2075-2085}
}

@inproceedings{wang2003adaptive,
  title={Adaptive monitoring for video surveillance},
  author={Wang, Jun and Yan, W-Q and Kankanhalli, Mohan S and Jain, Ramesh and Reinders, Marcel JT},
  booktitle={Fourth International Conference on Information, Communications and Signal Processing, 2003 and the Fourth Pacific Rim Conference on Multimedia. Proceedings of the 2003 Joint},
  volume={2},
  pages={1139--1143},
  year={2003},
  organization={IEEE}
}

@article{hall2022systematic,
  title={A systematic study of bias amplification},
  author={Hall, Melissa and van der Maaten, Laurens and Gustafson, Laura and Jones, Maxwell and Adcock, Aaron},
  journal={arXiv preprint arXiv:2201.11706},
  year={2022}
}

@inproceedings{kotek2023gender,
  title={Gender bias and stereotypes in large language models},
  author={Kotek, Hadas and Dockum, Rikker and Sun, David},
  booktitle={Proceedings of the ACM collective intelligence conference},
  pages={12--24},
  year={2023}
}

@article{navigli2023biases,
  title={Biases in large language models: origins, inventory, and discussion},
  author={Navigli, Roberto and Conia, Simone and Ross, Bj{\"o}rn},
  journal={ACM Journal of Data and Information Quality},
  volume={15},
  number={2},
  pages={1--21},
  year={2023},
  publisher={ACM New York, NY}
}

@incollection{ruggeri2023multi,
  title={A Multi-dimensional study on Bias in Vision-Language models},
  author={Ruggeri, Gabriele and Nozza, Debora and others},
  booktitle={Findings of the Association for Computational Linguistics: ACL 2023},
  year={2023},
  publisher={Association for Computational Linguistics}
}

@article{fraser2024examining,
  title={Examining gender and racial bias in large vision-language models using a novel dataset of parallel images},
  author={Fraser, Kathleen C and Kiritchenko, Svetlana},
  journal={arXiv preprint arXiv:2402.05779},
  year={2024}
}

@book{schwartz2022towards,
  title={Towards a standard for identifying and managing bias in artificial intelligence},
  author={Schwartz, Reva and Schwartz, Reva and Vassilev, Apostol and Greene, Kristen and Perine, Lori and Burt, Andrew and Hall, Patrick},
  volume={3},
  year={2022},
  publisher={US Department of Commerce, National Institute of Standards and Technology~…}
}

@inproceedings{lin2014microsoft,
  title={Microsoft coco: Common objects in context},
  author={Lin, Tsung-Yi and Maire, Michael and Belongie, Serge and Hays, James and Perona, Pietro and Ramanan, Deva and Doll{\'a}r, Piotr and Zitnick, C Lawrence},
  booktitle={Computer vision--ECCV 2014: 13th European conference, zurich, Switzerland, September 6-12, 2014, proceedings, part v 13},
  pages={740--755},
  year={2014},
  organization={Springer}
}

@inproceedings{panayotov2015librispeech,
  title={Librispeech: an asr corpus based on public domain audio books},
  author={Panayotov, Vassil and Chen, Guoguo and Povey, Daniel and Khudanpur, Sanjeev},
  booktitle={2015 IEEE international conference on acoustics, speech and signal processing (ICASSP)},
  pages={5206--5210},
  year={2015},
  organization={IEEE}
}

@inproceedings{gemmeke2017audio,
  title={Audio set: An ontology and human-labeled dataset for audio events},
  author={Gemmeke, Jort F and Ellis, Daniel PW and Freedman, Dylan and Jansen, Aren and Lawrence, Wade and Moore, R Channing and Plakal, Manoj and Ritter, Marvin},
  booktitle={2017 IEEE international conference on acoustics, speech and signal processing (ICASSP)},
  pages={776--780},
  year={2017},
  organization={IEEE}
}

@article{paszke2019pytorch,
  title={Pytorch: An imperative style, high-performance deep learning library},
  author={Paszke, A},
  journal={arXiv preprint arXiv:1912.01703},
  year={2019}
}

@misc{tensorflow2015-whitepaper,
title={ {TensorFlow}: Large-Scale Machine Learning on Heterogeneous Systems},
url={https://www.tensorflow.org/},
note={Software available from tensorflow.org},
author={
    Mart\'{\i}n~Abadi and
    Ashish~Agarwal and
    Paul~Barham and
    Eugene~Brevdo and
    Zhifeng~Chen and
    Craig~Citro and
    Greg~S.~Corrado and
    Andy~Davis and
    Jeffrey~Dean and
    Matthieu~Devin and
    Sanjay~Ghemawat and
    Ian~Goodfellow and
    Andrew~Harp and
    Geoffrey~Irving and
    Michael~Isard and
    Yangqing Jia and
    Rafal~Jozefowicz and
    Lukasz~Kaiser and
    Manjunath~Kudlur and
    Josh~Levenberg and
    Dandelion~Man\'{e} and
    Rajat~Monga and
    Sherry~Moore and
    Derek~Murray and
    Chris~Olah and
    Mike~Schuster and
    Jonathon~Shlens and
    Benoit~Steiner and
    Ilya~Sutskever and
    Kunal~Talwar and
    Paul~Tucker and
    Vincent~Vanhoucke and
    Vijay~Vasudevan and
    Fernanda~Vi\'{e}gas and
    Oriol~Vinyals and
    Pete~Warden and
    Martin~Wattenberg and
    Martin~Wicke and
    Yuan~Yu and
    Xiaoqiang~Zheng},
  year={2015},
}

@article{wolf2019huggingface,
  title={Huggingface's transformers: State-of-the-art natural language processing},
  author={Wolf, Thomas and Debut, Lysandre and Sanh, Victor and Chaumond, Julien and Delangue, Clement and Moi, Anthony and Cistac, Pierric and Rault, Tim and Louf, R{\'e}mi and Funtowicz, Morgan and others},
  journal={arXiv preprint arXiv:1910.03771},
  year={2019}
}

@book{sutton1998reinforcement,
  title={Reinforcement learning: An introduction},
  author={Sutton, Richard S and Barto, Andrew G and others},
  volume={1},
  number={1},
  year={1998},
  publisher={MIT press Cambridge}
}

@book{landau2011adaptive,
  title={Adaptive control: algorithms, analysis and applications},
  author={Landau, Ioan Dor{\'e} and Lozano, Rogelio and M'Saad, Mohammed and Karimi, Alireza},
  year={2011},
  publisher={Springer Science \& Business Media}
}

@article{hoi2021online,
  title={Online learning: A comprehensive survey},
  author={Hoi, Steven CH and Sahoo, Doyen and Lu, Jing and Zhao, Peilin},
  journal={Neurocomputing},
  volume={459},
  pages={249--289},
  year={2021},
  publisher={Elsevier}
}

@inproceedings{greff2022kubric,
  title={Kubric: A scalable dataset generator},
  author={Greff, Klaus and Belletti, Francois and Beyer, Lucas and Doersch, Carl and Du, Yilun and Duckworth, Daniel and Fleet, David J and Gnanapragasam, Dan and Golemo, Florian and Herrmann, Charles and others},
  booktitle={Proceedings of the IEEE/CVF conference on computer vision and pattern recognition},
  pages={3749--3761},
  year={2022}
}

@article{mittal2023orbit,
   author={Mittal, Mayank and Yu, Calvin and Yu, Qinxi and Liu, Jingzhou and Rudin, Nikita and Hoeller, David and Yuan, Jia Lin and Singh, Ritvik and Guo, Yunrong and Mazhar, Hammad and Mandlekar, Ajay and Babich, Buck and State, Gavriel and Hutter, Marco and Garg, Animesh},
   journal={IEEE Robotics and Automation Letters},
   title={Orbit: A Unified Simulation Framework for Interactive Robot Learning Environments},
   year={2023},
   volume={8},
   number={6},
   pages={3740-3747},
   doi={10.1109/LRA.2023.3270034}
}

@inproceedings{todorov2012mujoco,
  title={MuJoCo: A physics engine for model-based control},
  author={Todorov, Emanuel and Erez, Tom and Tassa, Yuval},
  booktitle={2012 IEEE/RSJ International Conference on Intelligent Robots and Systems},
  pages={5026--5033},
  year={2012},
  organization={IEEE},
  doi={10.1109/IROS.2012.6386109}
}

@article{li2023robustness,
  title={Robustness of visual perception system in progressive challenging weather scenarios},
  author={Li, Xingge and Zhang, Shufeng and Chen, Xun and Wang, Yashun and Fan, Zhengwei and Pang, Xiaofei and Hu, Jingwen},
  journal={Engineering Applications of Artificial Intelligence},
  volume={119},
  pages={105740},
  year={2023},
  publisher={Elsevier}
}

@inproceedings{caesar2020nuscenes,
  title={nuscenes: A multimodal dataset for autonomous driving},
  author={Caesar, Holger and Bankiti, Varun and Lang, Alex H and Vora, Sourabh and Liong, Venice Erin and Xu, Qiang and Krishnan, Anush and Pan, Yu and Baldan, Giancarlo and Beijbom, Oscar},
  booktitle={Proceedings of the IEEE/CVF conference on computer vision and pattern recognition},
  pages={11621--11631},
  year={2020}
}

@article{jin2023development,
  title={Development of robust detector using the weather deep generative model for outdoor monitoring system},
  author={Jin, Kyo-Hoon and Kang, Kyung-Su and Shin, Baek-Kyun and Kwon, June-Hyoung and Jang, Soo-Jin and Kim, Young-Bin and Ryu, Han-Guk},
  journal={Expert Systems with Applications},
  volume={234},
  pages={120984},
  year={2023},
  publisher={Elsevier}
}

@inproceedings{humphries2020impact,
  title={Impact of illuminance on object detection in industrial vision systems using neural networks},
  author={Humphries, Jacqueline and O'Brien, Kelly and O'Brien, Luke and Dunne, Nathan},
  booktitle={2020 2nd International Conference on Artificial Intelligence, Robotics and Control},
  pages={24--28},
  year={2020}
}

@article{van2021sustainable,
  title={Sustainable AI: AI for sustainability and the sustainability of AI},
  author={Van Wynsberghe, Aimee},
  journal={AI and Ethics},
  volume={1},
  number={3},
  pages={213--218},
  year={2021},
  publisher={Springer}
}

@article{duan2022survey,
  title={A survey of embodied ai: From simulators to research tasks},
  author={Duan, Jiafei and Yu, Samson and Tan, Hui Li and Zhu, Hongyuan and Tan, Cheston},
  journal={IEEE Transactions on Emerging Topics in Computational Intelligence},
  volume={6},
  number={2},
  pages={230--244},
  year={2022},
  publisher={IEEE}
}

@inproceedings{averly2023unified,
  title={Unified out-of-distribution detection: A model-specific perspective},
  author={Averly, Reza and Chao, Wei-Lun},
  booktitle={Proceedings of the IEEE/CVF International Conference on Computer Vision},
  pages={1453--1463},
  year={2023}
}

@article{yang2024generalized,
  title={Generalized out-of-distribution detection: A survey},
  author={Yang, Jingkang and Zhou, Kaiyang and Li, Yixuan and Liu, Ziwei},
  journal={International Journal of Computer Vision},
  volume={132},
  number={12},
  pages={5635--5662},
  year={2024},
  publisher={Springer}
}

@book{puterman2014markov,
  title={Markov decision processes: discrete stochastic dynamic programming},
  author={Puterman, Martin L},
  year={2014},
  publisher={John Wiley \& Sons}
}

@article{van2024big,
  title={Big AI: Cloud infrastructure dependence and the industrialisation of artificial intelligence},
  author={Van Der Vlist, Fernando and Helmond, Anne and Ferrari, Fabian},
  journal={Big Data \& Society},
  volume={11},
  number={1},
  pages={20539517241232630},
  year={2024},
  publisher={SAGE Publications Sage UK: London, England}
}

@software{openclip,
  author       = {Ilharco, Gabriel and
                  Wortsman, Mitchell and
                  Wightman, Ross and
                  Gordon, Cade and
                  Carlini, Nicholas and
                  Taori, Rohan and
                  Dave, Achal and
                  Shankar, Vaishaal and
                  Namkoong, Hongseok and
                  Miller, John and
                  Hajishirzi, Hannaneh and
                  Farhadi, Ali and
                  Schmidt, Ludwig},
  title        = {OpenCLIP},
  month        = {jul},
  year         = {2021},
  note         = {If you use this software, please cite it as below.},
  publisher    = {Zenodo},
  version      = {0.1},
  doi          = {10.5281/zenodo.5143773},
  url          = {https://doi.org/10.5281/zenodo.5143773}
}

@article{manion2024effect,
  title={The effect of pupil size on visual resolution},
  author={Manion, Garrett and Stokkermans, Thomas},
  journal={StatPearls},
  year={2024}
}

@article{webster2024designing,
  title={Designing for Sensory Adaptation: What You See Depends on What You’ve Been Looking at-Recommendations, Guidelines and Standards Should Reflect This},
  author={Webster, Michael A and Parthasarathy, Mohana Kuppuswamy and Zuley, Margarita L and Bandos, Andriy I and Whitehead, Lorne and Abbey, Craig K},
  journal={Policy Insights from the Behavioral and Brain Sciences},
  volume={11},
  number={1},
  pages={43--50},
  year={2024},
  publisher={Sage Publications Sage CA: Los Angeles, CA}
}

@article{gutschick2002should,
  title={Should you use a digital camera in your research?},
  author={Gutschick, Vincent},
  journal={Bulletin of the ecological society of America},
  volume={83},
  number={3},
  pages={176--180},
  year={2002},
  publisher={JSTOR}
}

@book{moore2012introduction,
  title={An introduction to the psychology of hearing},
  author={Moore, Brian CJ},
  year={2012},
  publisher={Brill}
}

@article{lanting2013mechanisms,
  title={Mechanisms of adaptation in human auditory cortex},
  author={Lanting, Cornelis P and Briley, Paul M and Sumner, Christian J and Krumbholz, Katrin},
  journal={Journal of neurophysiology},
  volume={110},
  number={4},
  pages={973--983},
  year={2013},
  publisher={American Physiological Society Bethesda, MD}
}

@article{shah2019design,
  title={Design approaches of MEMS microphones for enhanced performance},
  author={Shah, Muhammad Ali and Shah, Ibrar Ali and Lee, Duck-Gyu and Hur, Shin},
  journal={Journal of sensors},
  volume={2019},
  number={1},
  pages={9294528},
  year={2019},
  publisher={Wiley Online Library}
}

@article{van2016haptic,
  title={Haptic adaptation to slant: No transfer between exploration modes},
  author={Van Dam, Loes CJ and Plaisier, Myrthe A and Glowania, Catharina and Ernst, Marc O},
  journal={Scientific Reports},
  volume={6},
  number={1},
  pages={34412},
  year={2016},
  publisher={Nature Publishing Group UK London}
}

@article{jiang2025can,
  title={How Can Haptic Feedback Assist People with Blind and Low Vision (BLV): A Systematic Literature Review},
  author={Jiang, Chutian and Kuang, Emily and Fan, Mingming},
  journal={ACM Transactions on Accessible Computing},
  volume={18},
  number={1},
  pages={1--57},
  year={2025},
  publisher={ACM New York, NY}
}

@article{guo2024active,
  title={Active electronic skin: an interface towards ambient haptic feedback on physical surfaces},
  author={Guo, Yuan and Wang, Yun and Tong, Qianqian and Shan, Boxue and He, Liwen and Zhang, Yuru and Wang, Dangxiao},
  journal={Npj Flexible Electronics},
  volume={8},
  number={1},
  pages={25},
  year={2024},
  publisher={Nature Publishing Group UK London}
}

@article{johansson2009coding,
  title={Coding and use of tactile signals from the fingertips in object manipulation tasks},
  author={Johansson, Roland S and Flanagan, J Randall},
  journal={Nature Reviews Neuroscience},
  volume={10},
  number={5},
  pages={345--359},
  year={2009},
  publisher={Nature Publishing Group UK London}
}

@article{miyazaki2014image,
  title={The image-scratch paradigm: a new paradigm for evaluating infants' motivated gaze control},
  author={Miyazaki, Michiko and Takahashi, Hideyuki and Rolf, Matthias and Okada, Hiroyuki and Omori, Takashi},
  journal={Scientific reports},
  volume={4},
  number={1},
  pages={5498},
  year={2014},
  publisher={Nature Publishing Group UK London}
}

@article{yu2024active,
  title={Active Gaze Behavior Boosts Self-Supervised Object Learning},
  author={Yu, Zhengyang and Aubret, Arthur and Raabe, Marcel C and Yang, Jane and Yu, Chen and Triesch, Jochen},
  journal={arXiv preprint arXiv:2411.01969},
  year={2024}
}

@inproceedings{castro2024artificial,
  title={Artificial Intelligence for All: Challenges and Harnessing Opportunities in AI Democratization},
  author={Castro, Keylin Alessandra and Siwady, Joselyn Alvarado and Castillo, Erika and Alonzo, Alberto and Cardona, Manuel and Perdomo, Mar{\'\i}a Elena},
  booktitle={2024 IEEE International Conference on Machine Learning and Applied Network Technologies (ICMLANT)},
  pages={143--148},
  year={2024},
  organization={IEEE}
}



\end{document}